\documentclass[10pt,journal,compsoc]{IEEEtran}
\ifCLASSOPTIONcompsoc
  \usepackage[nocompress]{cite}
\else
  \usepackage{cite}
\fi
\ifCLASSINFOpdf
\else
\fi

\usepackage{threeparttable}

\usepackage{arydshln} 
\usepackage{booktabs}
\usepackage{multirow}
\usepackage{pifont}
\usepackage{color}
\usepackage{amssymb}
\usepackage{amsthm}
\usepackage{graphicx}
\usepackage{color}
\usepackage{dsfont}
\usepackage{algorithm}
\usepackage{algorithmicx}
\usepackage{algpseudocode}
\usepackage{tabularx}
\usepackage{multirow}
\usepackage{amsthm,amsmath,amssymb} 
\usepackage{mathrsfs}
\usepackage{array}
\usepackage{booktabs}  %
\usepackage{slashed}
\usepackage{scalerel}
\usepackage[top=0.4in,bottom=0.4in,left=0.4in,textwidth=7.7in]{geometry}
\usepackage{xspace}
\makeatletter 
\DeclareRobustCommand\onedot{\futurelet\@let@token\@onedot}
\def\@onedot{\ifx\@let@token.\else.\null\fi\xspace}

\def\eg{\emph{e.g}\onedot} 
\def\ie{\emph{i.e}\onedot} 
 
\def\etc{\emph{etc}\onedot} 
 
\def\etal{\emph{et al}\onedot}
\makeatother

\usepackage[pagebackref=false,breaklinks=true,linkcolor=red,anchorcolor=blue, citecolor=green,colorlinks,bookmarks=false]{hyperref}
\begin{document}

\title{Unbiased Scene Graph Generation via \\ Two-stage Causal Modeling}

\author{
Shuzhou Sun, Shuaifeng Zhi, Qing Liao, Janne Heikkilä, Li Liu
\IEEEcompsocitemizethanks{
\IEEEcompsocthanksitem This work was partially supported by the National Key Research and Development Program of China No. 2021YFB3100800, the Academy of Finland under grant 331883, Infotech Project FRAGES, and the National Natural Science Foundation of China under Grant 61872379, 62022091, 62201603 and 62201588. 
\IEEEcompsocthanksitem Li Liu (dreamliu2010@gmail.com) and shuaifeng Zhi are with the College of Electronic Science, National University of Defense Technology (NUDT), Changsha, Hunan, China. Li Liu is also with the CMVS at the University of Oulu, Finland. \protect \\
 Li Liu is the corresponding author.
\IEEEcompsocthanksitem Qing Liao is with the Department of Computer Science and Technology, Harbin Institute of Technology, Shenzhen, China, and also with the Peng Cheng Laboratory, Shenzhen, China, 518055. Shuzhou Sun and Janne Heikkilä are with the Center for Machine Vision and Signal Analysis,  University of Oulu, Finland.}
}

\markboth{IEEE Transactions on Pattern Analysis and Machine Intelligence}%
{Sun \MakeLowercase{\textit{et al.}}: USGG}

\IEEEtitleabstractindextext{%
\begin{abstract}
Despite the impressive performance of recent unbiased Scene Graph Generation (SGG) methods, the current debiasing literature mainly focuses on the long-tailed distribution problem, whereas it overlooks another source of bias, \ie, semantic confusion, which makes the SGG model prone to yield false predictions for similar relationships. In this paper, we explore a debiasing procedure for the SGG task leveraging causal inference. Our central insight is that the Sparse Mechanism Shift (SMS) in causality allows independent intervention on multiple biases, thereby potentially preserving head category performance while pursuing the prediction of high-informative tail relationships. However, the noisy datasets lead to unobserved confounders for the SGG task, and thus the constructed causal models are always causal-insufficient to benefit from SMS. To remedy this, we propose Two-stage Causal Modeling (TsCM) for the SGG task, which takes the long-tailed distribution and semantic confusion as confounders to the Structural Causal Model (SCM) and then decouples the causal intervention into two stages. The first stage is causal representation learning, where we use a novel Population Loss (P-Loss) to intervene in the semantic confusion confounder. The second stage introduces the Adaptive Logit Adjustment (AL-Adjustment) to eliminate the long-tailed distribution confounder to complete causal calibration learning. These two stages are model agnostic and thus can be used in any SGG model that seeks unbiased predictions. Comprehensive experiments conducted on the popular SGG backbones and benchmarks show that our TsCM can achieve state-of-the-art performance in terms of mean recall rate. Furthermore, TsCM can maintain a higher recall rate than other debiasing methods, which indicates that our method can achieve a better tradeoff between head and tail relationships. 
\end{abstract}

}

\maketitle

\IEEEdisplaynontitleabstractindextext
\IEEEpeerreviewmaketitle

\IEEEraisesectionheading{\section{Introduction}\label{sec:introduction}}

Scene Graph Generation (SGG), first proposed by Scherrer \etal.\cite{SceneGraph}, is an emerging, critical, and challenging intermediate scene-understanding task and has received increasing attention, especially during the past few years \cite{SGGSurvey,add_sgg_1}, due to its potential to be a bridge between computer vision and natural language processing. SGG aims to generate a structured representation of a scene that jointly describes objects and their attributes, as well as their pairwise relationships, and is typically formulated as a set of \textit {$<$subject, relationship, object$>$} triplets. Such representations can provide a deep understanding of a scene, and thus SGG has been employed for many downstream tasks, such as image-text retrieval\cite{SceneGraph,retrieval}, visual question answering \cite{VQA1,VQA2}, visual captioning \cite{imagecaptioning1,imagecaptioning2}, \etc.

\begin{figure*}
	\footnotesize\centering
	\centerline{\includegraphics[width=1\linewidth]{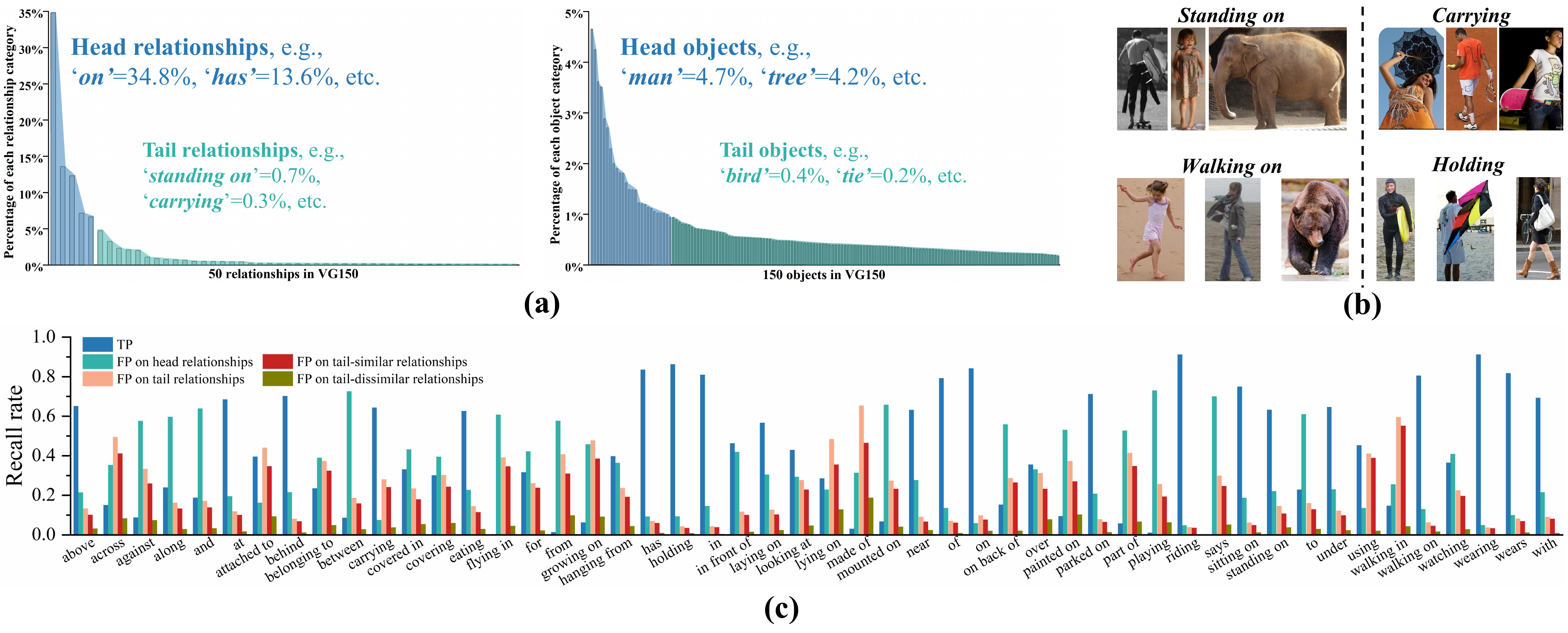}}
        \caption{The motivations of TsCM. (a) illustrates the long-tailed distribution bias. (b) shows the semantic confusion bias. (c) reports the True Predictions (TP), False Predictions (FP) on head relationships, and FP on tail relationships (further divided into two cases depending on whether the predictions are tail-similar relationships or not) of the MotifsNet \cite{Neuralmotifs} framework. Formally, in this paper, head relationships refer to \textit {on}, \textit {has}, \textit {in}, \textit {of}, and \textit {wearing}, since they account for more than 50\%, and the rest are tail relationships, but note often different grouping criteria were also adopted in the literature \cite{FGPL,NICE,LSKD,CAME}. For a given category, similar and dissimilar relationships are those found within and outside its population, respectively. A formal introduction to the concept of population is provided in Section \ref{sec3.2.2}.}
	\label{motivation}
\end{figure*}

While early SGG work has made significant progress \cite{Neuralmotifs,SceneGraph,Earlywork}, which, however, as discussed in \cite{VCtree,KERN}, tends to generate biased predictions, \ie, informative fine-grained relationships (\eg, \textit {standing on}) are predicted as less informative coarse-grained relationships (\eg, \textit {on}) due to the long-tailed distribution problem. As an example, we consider the distribution of the relationships in VG150 \cite{VG150}, a popular benchmark in the SGG task, which, as shown in Fig.~\ref{motivation} (a), clearly suffers from severe long-tailed distribution problems. The SGG model, naturally, cannot learn to represent the features of the head and tail relationships simultaneously from the skewed distribution and, hence, easily yields False Predictions (FP) on head relationships (see Fig.~\ref{motivation} (c)). 

For the above biased predictions, many debiasing methods \cite{TDE,EBMloss,GCL,PPDL,add_sgg_2} have been proposed to overcome this problem. Unlike earlier work, the primary goal of debiasing methods, however, is to pursue the unbiased scene graphs. Existing debiasing methods can be roughly categorized into four groups: 1) \textit {Resampling methods} \cite{SegG,TransRwt} upsample the tail relationships and/or downsample the head relationships to rebalance the training data distribution. 2) \textit {Reweighting methods} \cite{Cogtree,PPDL,EBMloss,add_longtail_2,add_longtail_3} revise the contribution of different relationships during training, for instance, weighting the prediction loss to strengthen the model's representation ability to the tail categories. 3) \textit {Adjustment methods} \cite{TDE,DLFE,PKO,post_hoc1} modify the learned biased model to obtain unbiased predictions, for example, by adjusting the output logits to increase the likelihood of more informative fine-grained relationships. 4) \textit {Hybrid methods} \cite{BPLSA,HML,RTPB} combine some/all of the above methods. Although debiasing research is rather active in the SGG community, the above methods often fall short in preserving head category performance while pursuing the prediction of informative tail relationships \cite{SGGSurvey,TDE}. More importantly, the current debiasing methods mainly focus on a single bias, \ie, the long-tailed distribution problem, whereas it overlooks other biases. 

Unlike existing work that focuses on a single bias, we reveal the fact that there are multiple biases for the SGG task in this paper. This stems from our observation that some of the False Predictions (FP) are clearly not caused by the long-tailed distribution bias, \eg, FP on tail relationships (see Fig.~\ref{motivation} (c)). We therefore argue that there are other biases that have not yet been observed and explored with current debiasing methods. Cognitive psychology \cite{Cognitive} and studies on the human visual system \cite{humanvisualsystem} suggest that humans struggle to distinguish similar objects. Inspired by this fact, we hypothesize that the source of the bias of FP on tail relationships is semantic confusion, which refers to two relationships sharing similar semantic information. For instance, as shown in Fig.~\ref{motivation} (b), both \textit {carrying} and \textit {holding} are semantic concepts composed of a people and objects in his/her hands. To demonstrate our premise, as shown in Fig.~\ref{motivation} (c), we additionally split FP on tail relationships into FP on tail-similar relationships and FP on tail-dissimilar relationships. As expected, most of the FP on tail relationships occur in tail-similar relationships. This suggests that SGG models, like humans, have difficulties in distinguishing similar relationships. As a result, we take semantic confusion as the second bias. 

For the multiple biases in the SGG task, we seek causal inference \cite{pearl2009causality,causalGeneralization}, an inference procedure that achieves impressive performance in statistics, econometrics, and epidemiology, which has also attracted significant attention in the deep learning community in recent years. Our central insight is that the Sparse Mechanism Shift (SMS) \cite{bengio2019meta,Bengio2021Toward} in causal inference allows independent intervention on multiple biases, thereby potentially preserving head category performance while pursuing higher performance in fine-grained tail relationships. Inspired by Pearl Causal Hierarchy (PCH) \cite{PCH}, in particular its highest layer, counterfactual, we pose two questions: 1) What happens if there is no semantic confusion between any two relationships in the observed data? 2) What happens if the distribution of relationships in the observed data is balanced? To answer these two counterfactual questions, we first build Structural Causal Models (SCM) \cite{pearlCBM,CausalFairness}, a causal modeling method that can support counterfactual inference, based on two observed biases as confounders. In practice, unfortunately, not all confounders for the SGG task can be observed, which means that the built SCM is causal-insufficient (see Section \ref{sec3.1} for a detailed analysis). Causal-insufficient assumption will invalidate the SMS hypothesis because the variables of the SCM are entangled in this case. Put another way, when we use existing causal intervention methods to overcome the observed biases, unobserved biases could be disturbed and bring about unwanted consequences. To allow SCM with causal-insufficient assumption to also benefit from the SMS hypothesis, we decouple the causal interventions into two stages and, on this basis, propose a novel causal modeling method, Two-stage Causal Modeling (TsCM), tailored for the SGG task.

Our TsCM consists of two stages: 1) Stage 1, causal representation learning, where despite the causal-insufficient assumption of the built SCM, we find that similarity relationships have inherently sparse properties (see Section \ref{sec3.2}), and, hence, sparse perturbations and independent interventions on semantic confusion bias are attainable. To achieve this, we proposed the Population Loss (P-Loss), which intervenes in the model training process to increase the prediction gap between similar relationships, allowing the trained model to obtain the causal representation that can eliminate the semantic confusion bias. As a result, this stage disentangles the confusion bias from the variables of the built SCM, thereby getting a disentangled factorization. 2) Stage 2, causal calibration learning, where thanks to the disentangled factorization obtained in stage 1, we calibrate the model's causal representation to remove the long-tailed distribution bias. Specifically, this is achieved by our proposed Adaptive Logit Adjustment (AL-Adjustment), which can adaptively learn a set of adjustment factors from the observed data for sparse perturbations and independent interventions.

In summary, the contributions of our work are three-fold:
\begin{itemize}
\item We thoroughly analyze the sources of bias in the biased SGG model and experimentally verify the bias, \ie, semantic confusion bias, ignored by current debiasing methods.

\item We propose a new causal modeling framework, Two-stage Causal Modeling (TsCM), to disentangle the multiple biases from the biased SGG model. Our TsCM decouples the causal intervention into two stages. Stage 1 leverages the proposed P-Loss to remove the semantic confusion bias and obtain a disentangled factorization even in the case of insufficient causality, thereby providing the causal representation that can distinguish similar relationships. Stage 2 further calibrates the causal representation to eliminate the long-tailed distribution bias by using the proposed AL-Adjustment.

\item Comprehensive experiments on various SGG backbones and the popular benchmark demonstrate the state-of-the-art mean recall rate of the proposed TsCM. Furthermore, our TsCM can maintain a higher recall rate than other debiasing methods, achieving a better tradeoff between head and tail relationships.
\end{itemize}

\section{Related works}
\subsection{Scene Graph Generation}
SGG produces a structured representation of the scene by assigning appropriate relationships to object pairs and enables a more comprehensive understanding of the scene for intelligent agents \cite{SceneGraph}. Most early works struggled with employing advanced network structures, \eg, Convolutional Neural Network, Recurrent Neural Network, Graph Neural Network, for better feature extraction and representation \cite{SGGSurvey,add_sgg_3}. Despite continuous improvements in the recall rate, these methods fall into the trap of biased prediction, \ie, informative fine-grained relationships are predicted as less informative coarse-grained relationships. As a result, debiasing methods have attracted unprecedented attention in the SGG community in recent years. To keep focus, here we mainly review the debiasing methods for the SGG task. Existing debiasing methods can be roughly categorized into four groups as follows.

\textit {Resampling methods} downsample the head category relationships and/or upsample the tail ones to balance the training data distribution, and often the prior knowledge, \eg, language prior, is taken into account, too. For instance, instead of relying on box-level annotations, SegG \cite{SegG} argues that pixel-level grounding would naturally be more valuable and, hence, create segmentation annotations for the SGG dataset with the help of auxiliary datasets. Recently, TransRwt \cite{TransRwt} rectified the skewed distribution by creating an enhanced dataset using Internal Transfer and External Transfer, the former for transferring the coarse-grained relationships to the fine-grained ones and the latter for re-labeling the relationships that are missing annotations. However, resampling methods may lead to overfitting (oversampling) or information loss (undersampling) by altering relationship category sample distributions.

\textit {Reweighting methods} design debiasing loss functions to make the model pay more attention to the tail category relationships or to create advanced networks to improve the representation ability of these relationships. Some works in this group begin by extracting prior knowledge from biased distributions, \eg, cognitive structure in CogTree \cite{Cogtree}, predicate lattice in FGPL \cite{FGPL}, relationship probability distribution in PPDL \cite{PPDL}, \etc, and then combine the proposed debiasing loss functions to supervise the model training. Besides, GCL \cite{GCL} presents a Stacked Hybrid-Attention network to achieve intra-modal refinement and intermodal interaction and then enhances the representation ability of tail relationships. Nonetheless, reweighting methods may result in an imbalanced focus on relationship categories and suboptimal, unstable performance due to manual or heuristic weight adjustments.  

\textit {Adjustment methods} adjust the output of the biased trained model to obtain unbiased predictions. The adjustment procedure can be based on prior knowledge. For example, Logit-reweight \cite{logitadjustment} uses label frequencies to adjust the logit outputs by the biased model. DLFE \cite{DLFE} considers the SGG task as a Learning from Positive and Unlabeled data (PU learning) problem, where a target PU dataset contains only positive examples and unlabeled data. However, its prior knowledge, \ie, label frequencies, is obtained iteratively during the training process by the proposed Dynamic Label Frequency Estimation method. Furthermore, adjustment procedures can also be modeled by causal inference. For instance, TDE \cite{TDE} first builds a causal graph for the SGG task and then draws counterfactual causality from the trained model to infer the effect from the negative bias. Note that adjustment methods will increase computational complexity with post-training output adjustments and may cause a decline in other relationship category performances.

\textit {Hybrid methods} combine some/all of the above techniques. HML \cite{HML} and CAME \cite{CAME} first divide the long-tailed distribution into some balanced subsets. HML \cite{HML} then trains the model with coarse-grained relationships and finally learns the fine-grained categories. While CAME \cite{CAME} then proposes to use a mixture of experts to handle different subsets. RTPB \cite{RTPB} enhances the impact of tail relationships on the training process based on prior bias and designs a contextual encoding backbone network to improve feature extraction capabilities. However, hybrid methods Increase implementation complexity, more challenging parameter tuning, higher computational costs, and potential performance instability due to the interplay of combined methods.

Despite achieving impressive results, the above debiasing methods focus almost exclusively on a single bias, \ie, long-tailed distribution bias, which clearly, makes complete debiasing impossible. Moreover, these methods sacrifice head relationships in pursuit of tail category performance. Differently, our method considers multiple biases and removes them using the causal inference technique. Our causal model TsCM consists of two stages covering both the reweighting and adjustment approaches. Thanks to the SMS mechanism, the two stages in our method independently intervene in different biases. In contrast, the different stages in existing hybrid methods only intervene in the same bias. 

\subsection{Causal Inference}

Causal analysis has achieved encouraging performance in health, social, behavioral sciences, \etc, and it has also attracted increasing attention in deep learning community in recent years, such as scene graph generation \cite{TDE}, out-of-distribution generalization \cite{causal_ood}, and salient object detection \cite{causal_detection}. Compared with deep learning models, the causal inference approaches can eliminate the influence of biases/confounders when making predictions \cite{pearl2003statistics,pearl2009causality}. A typical causal inference paradigm usually starts with establishing a graphical model, \eg, the Structural Causal Model (SCM) \cite{pearlCBM,CausalFairness,StatisticaltoCausal}, which models the dependencies of causal variables. It then intervenes in (\eg, \textit {do}-interventions \cite{pearlCBM,pearl2003statistics}) these variables to pursue causal inference of interest. The models can therefore be generalized to different distributions. 

It should be emphasized that the above interventions can be achieved because the causal variables satisfy the principle of sparse perturbation and independent intervention, which is the cornerstone of causal inference. The independent intervention principle in causality emphasizes that the conditional distribution of each causal variable, given its causes (\ie, its mechanism), does not inform or influence the other mechanisms. The sparse perturbation principle refers to small distribution changes that tend to manifest themselves in a sparse or local way \cite{bengio2019meta,SparseAhuja}. The sparse principle is extended by independence, which can be seen as a consequence of independence, too \cite{StatisticaltoCausal,Bengio2021Toward}. Benefiting from the independent intervention principle, Scherrer \etal.\cite{ICM} decompose the causal mechanism into modules with knowledge, which, different from monolithic models where full knowledge will be learned directly, enables adaptation to distribution shifts by only updating a subset of parameters. Thanks to the sparse perturbation principle, Ahuja \etal. \cite{SparseAhuja} achieve weakly supervised representation learning by perturbing the causal mechanism sparsely. Inspired by the above work and the multiple confounders in the SGG task, the model proposed in this paper removes these confounders independently and sparsely, which allows our model to preserve the performance of the head categories while pursuing debiasing.

\begin{figure*}
	\footnotesize\centering
	\centerline{\includegraphics[width=0.98\linewidth]{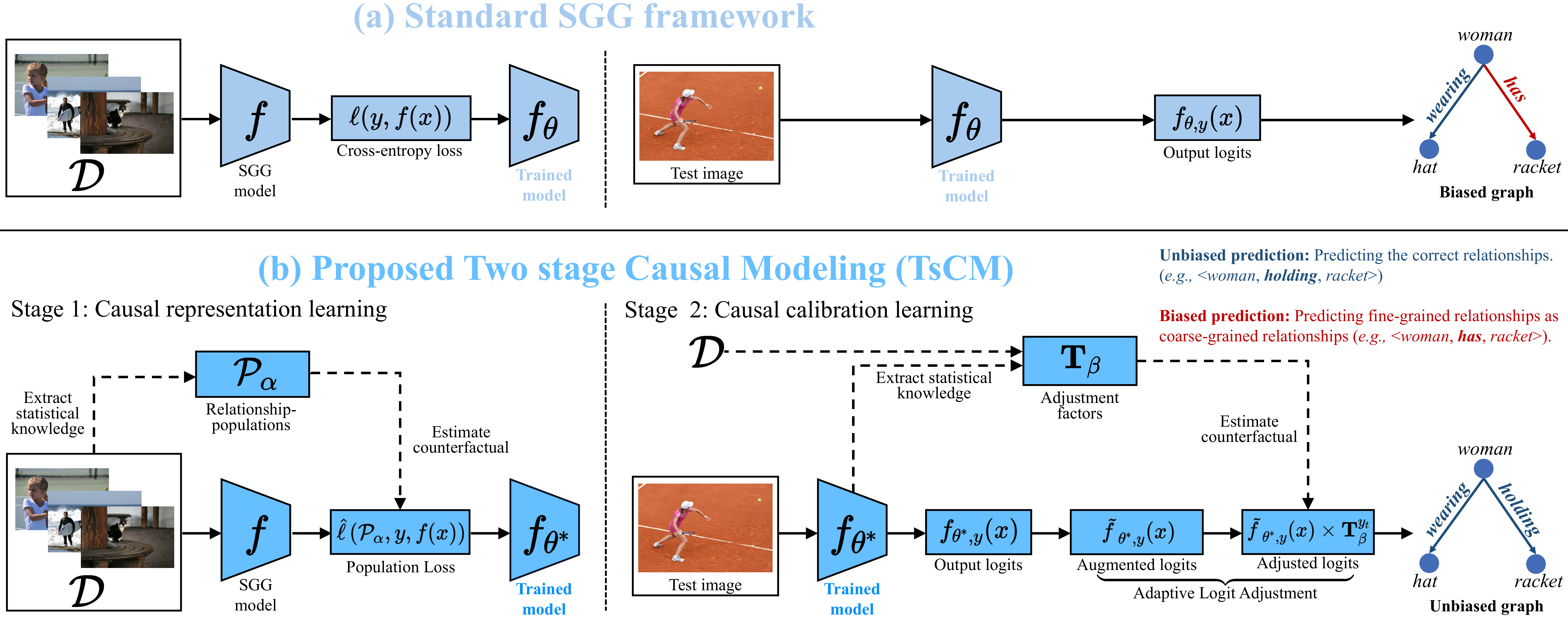}}
    \caption{The illustrations of the standard framework and our proposed pipeline. The standard SGG framework is supervised by cross-entropy loss. While TsCM consists of two stages: Stage 1 learns causal representation learning that can better distinguish semantic confusion as well as disentangle the causal-insufficient SCM. Specifically, we achieve this through the proposed Population Loss. Stage 2 leverages the proposed Adaptive Logit Adjustment to calibrate the entangled factorization from the previous stage. As a result, the adjusted logits can avoid the SSG model biases towards the head relationships.}
    \label{framework}
\end{figure*}

\section{Methods}
\label{sec3}

\subsection{Overview} 
\label{overview}
The primary goal of SGG is to model the objects existing in the scene and their pairwise relationships. Most existing works first detect the objects (\eg, ``man'', ``horse'' ) in the scene with an object detector and then recognize their pairwise relationships (\eg, ``riding'', ``standing on'') with a relationship classifier. The object detector extracts information about objects, like their bounding boxes, categories, and features. Then the relationship classifier predicts relationships for each pair of objects. Simply, a scene graph is a set of \emph{visual triples} in the formulation of $<$\emph{subject}, \emph{relationship}, \emph{object}$>$. Formally, let $\mathcal{D}=\{(\mathbf{x}_i, \mathbf{y}_i)\}_{i=1}^{N_{\mathcal{D}}}$ denote the observed data with $N_{\mathcal{D}}$ samples, where $\mathbf{x}_i$ is $i$-th image and $\mathbf{y}_i \in \mathbb{R}^{N_i \times \emph{K}}$ is $N_i$ relationships in this sample, $\mathbf{y}_{ij}$ is $\emph{K}$ dimension one-hot vector denoting the label of $j$-th relationship in $\mathbf{x}_i$. We therefore need to label the dataset $\mathcal{D}$ with visual triplets $\{<(\mathbf{o}_i^\textrm{sub},\mathbf{b}_i^\textrm{sub}), \mathbf{y}_i,(\mathbf{o}_i^\textrm{obj},\mathbf{b}_i^\textrm{obj})>\}_{i=1}^{N_{\mathcal{D}}}$ to support model training, where $\mathbf{o}_i^\textrm{sub}$, $\mathbf{o}_i^\textrm{obj}$ $\in \mathbb{R}^{N_i \times \emph{C}}$ and $\mathbf{b}_i^\textrm{sub}$, $\mathbf{b}_i^\textrm{obj}$ $\in \mathbb{R}^{N_i \times 4}$, $\mathbf{o}_{ij}$ and $\mathbf{b}_{ij}$ denoting the category  and bounding box information of subject or object of $j$-th relationship in $\mathbf{x}_i$ respectively. \emph{C} and \emph{K} are the numbers of categories of objects and relationships in the observed data, respectively. Although labeling the visual triples is very costly, early efforts have contributed a few benchmarks to the SGG community, such as Visual Genome \cite{VG}, Scene Graph \cite{SceneGraph}, and Open Images V4 \cite{OpenImagesV4}. However, SGG models trained on these datasets typically suffer from two challenges: (1) Semantic confusion, and (2) Long-tailed distribution.

In this work, we address the above two challenges from the perspective of causal inference. Specifically, in Section \ref{sec3.1}, we firstly consider the aforementioned two  challenges, \ie, semantic confusion and long-tailed distribution, as confounders for the standard SGG framework (see Fig.~\ref{framework} (a)). Therefore, our method leverages the data-level confounders to model the causality for the SGG task. Compared to model-level confounders \cite{TDE}, our approach is model-agnostic, \ie, transferable to arbitrary SGG models. We then propose the Population Loss in Section \ref{sec3.2} to remove the semantic confusion confounder and get a disentangled factorization for the causal model (see Stage 1 in Fig.~\ref{framework} (b)). Next, in Section \ref{sec3.3}, we propose AL-Adjustment to remove the long-tailed distribution confounder to obtain unbiased predictions (see Stage 2 in Fig.~\ref{framework}). Finally, in Section \ref{sec3.4}, we show that our method is Fisher-consistent and highlight the differences from existing statistical-based approaches.

\subsection{Modeling structural causal model}
\label{sec3.1}
One can use a variety of frameworks to model the causality of their system of interest, such as Causal Graphical Models (CGM), Structural Causal Models (SCM), and Potential Outcomes (PO) \cite{pearlCBM,CausalFairness,pearl2009causality,StatisticaltoCausal}. The causality modeling ability of CGM is limited since it cannot support counterfactual inference. PO is active in the system with binary treatment variables, but it is awkward when dealing with special treatment and outcome variables. Considering the limitations of CGM and PO, in this work, we model the causality using SCM, a structural method that contains variables, structural functions, and distributions over the variables (see Definition 1).

\noindent \textbf{Definition 1} (Structural Causal Model (SCM) \cite{pearlCBM,CausalFairness}). \textit { A structural causal model $(S C M)$ $\mathcal{M}$ is a 4-tuple $\langle \mathcal{V}, \mathcal{U}, \mathcal{F}, P(\mathcal{U})\rangle$, where $\mathcal{U} = \{ U_1, U_2, \cdot\cdot\cdot , U_n \}$ is a set of exogenous variables; $\mathcal{V} = \{ V_1, V_2, \cdot\cdot\cdot , V_n \}$ is a set of endogenous (observed) variables; $\mathcal{F}=\{ F_1, F_2, \cdot\cdot\cdot , F_n \}$ is the set of structural functions determining $\mathcal{V}$; $P(\mathcal{U})$ is a distribution over the exogenous variables.} 

\noindent \textbf{Definition 2} (Submodel \cite{pearlCBM,CausalFairness}). \textit { For the SCM $\mathcal{M}$, let $\widetilde{\mathcal{V}}$ be a set of variables in $\mathcal{V}$, and $\widetilde{v}$ a particular value of $\widetilde{\mathcal{V}}$. A submodel $\mathcal{M}_{\widetilde{v}}$ (of $\mathcal{M}$ ) is a 4-tuple: $\mathcal{M}_{\widetilde{v}}=\left\langle \mathcal{V}, \mathcal{U}, \mathcal{F}_{\widetilde{v}}, P(\mathcal{U})\right\rangle$, where $\mathcal{F}_{\widetilde{v}}=\{F_i: V_i \notin \widetilde{\mathcal{V}}\} \cup\{\widetilde{\mathcal{V}} \leftarrow {\widetilde{v}} $ \}, and all other components are preserved from $\mathcal{M}$.}

\begin{figure}
	\footnotesize\centering
	\centerline{\includegraphics[width=1\linewidth]{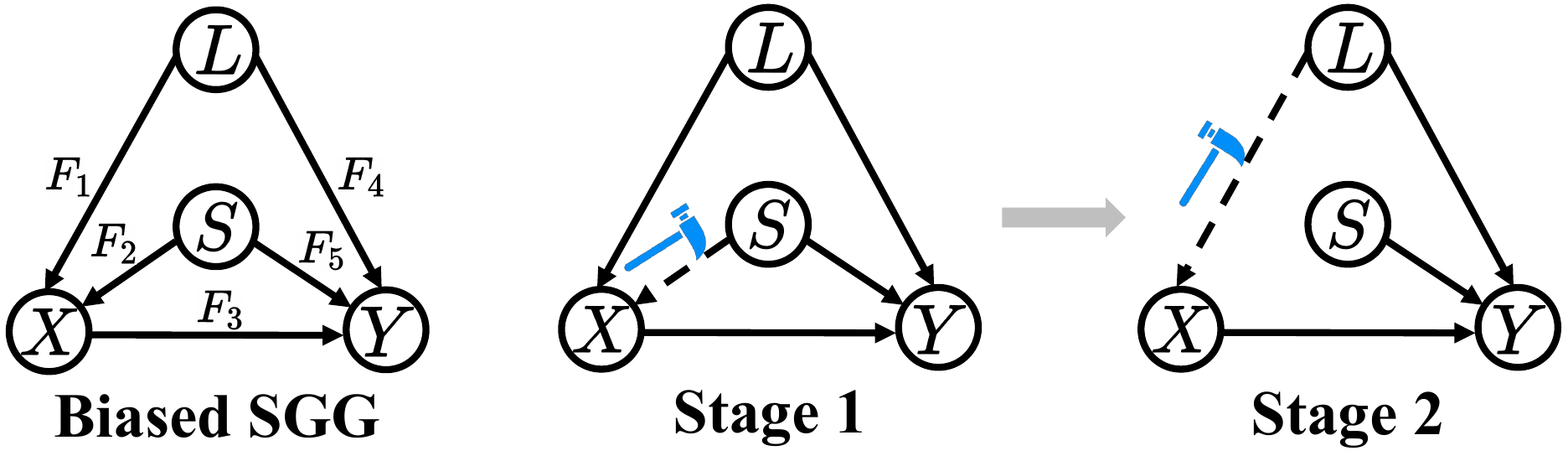}}
        \vspace{-0.2cm}
        \caption{The proposed Causal Graphical Model (SCM) for SGG.  $S$ and $L$ are data-level confounders, \ie, the semantic confusion confounder and the long-tailed distribution confounder. $X$ and $Y$ are the input image and the predicted relationships, respectively.}
	\label{SCM}
        \vspace{-0.3cm}
\end{figure}

Endogenous variables are the fundamental elements of an SCM. However, determining variable $\mathcal{V}$ in the SGG task is very challenging because its inputs, \ie, images, differ greatly from the structured units in traditional causal discovery and reasoning tasks \cite{CausalFairness,StatisticaltoCausal}. Inspired by FP on head/tail relationships in Fig.~\ref{motivation}, in this paper, we propose a model-agnostic data-level variable that takes semantic confusion and long-tailed distribution as the confounders. As a result, the induced submodel $\mathcal{M}_{\widetilde{v}}$ in our work is $\left\langle \mathcal{V}, \mathcal{U}, \mathcal{F}_{\widetilde{v}}, P(\mathcal{U})\right\rangle$ (see Definition 2). Where $\mathcal{V}=\{X, Y, S, L\}$, $X$ is input (images in SGG task), $Y$ is output (relationships), $S$ is the semantic confusion confounder, $L$ is the long-tailed distribution confounder; $\mathcal{U}=\left\{U_X, U_Y, U_S, U_L\right\}$; $\mathcal{F}_{\widetilde{v}}=\{F_1,F_2,F_3,F_4,F_5 \}$; $P(U_X, U_Y, U_S, U_L)$ is the distribution over the exogenous variables. The SCM is shown in Fig.~\ref{SCM} (Biased SGG), and thus the structural equations are:
\begin{equation}
\begin{aligned}
&S = P(S) ,\\
&L = P(L) ,\\
&X = F_1(L, P(L))+F_2(S, P(S)) ,\\
&Y = F_3(X, P(X))+F_4(L, P(L))+F_5(S, P(S)),
\end{aligned}
\end{equation}
Intuitively, we can directly use interventions to remove the confounders in SCM (see Definition 3). These interventions, however, do not update $P(\mathcal{U})$, and thus the intervened results are noisy causal effects in most cases \cite{pearl2009causality,CausalFairness}.

\noindent \textbf{Definition 3} (Interventions in SCM \cite{pearlCBM,StatisticaltoCausal}). \textit { An intervention $d o\left(V_i:=v'\right)$ in an SCM $\mathcal{M}$ is modeled by replacing the $i$-th structural equation by $V_i:=v'$, where $v'$ is a $V_i$-independent value.}

\noindent \textbf{Definition 4} (Counterfactual in SCM \cite{pearlCBM,StatisticaltoCausal}).\textit { A counterfactual in an SCM $\mathcal{M}$ is modeled by replacing the $i$-th structural equation by $V_i:=v'$ and update the $P(\mathcal{U})$, where $v'$ shares the same meaning as it does in Definition 3. The above counterfactual intervention induces the submodel $\mathcal{M}^{{V_i}}$. }

\noindent \textbf{Assumption 1} (Causal-insufficient). \textit {The exogenous variable $\mathcal{U}$ in $\mathcal{M}$ satisfies that: $P(U_1, \ldots, U_n) \neq P(U_1) \times P(U_2) \times \cdots \times P(U_n)$. }

Fortunately, counterfactual, the highest-level reasoning procedure of cognitive ability \cite{pearl2009causality}, overcomes this limitation by imagining pre/post-intervention results (see Definition 4). Note that counterfactual is unfalsifiable since its imaginary results cannot be observed. However, significant designs (\eg, average treatment effect) in statistics, econometrics, and epidemiology can estimate the counterfactuals and are proven effective. The principle difference between intervention and counterfactual is that the latter updates $P(\mathcal{U})$ when manipulating the structural equations \cite{CausalFairness}. Thus, one can partially seek the intervention technique to calculate the counterfactuals. Inspired by the above facts, we therefore leverage the counterfactual inference to eliminate the semantic confusion confounder $S$ and long-tailed distribution confounder $L$ to obtain an unbiased SCM $\mathcal{M}_{\widetilde{v}}^{{S},{L}}$ for the SGG task. The counterfactuals results can be calculated as:
\begin{equation}
\begin{aligned}
\mathbb{E}[Y \mid X, d o\left(S:= s\right)&, d o\left(L:=l\right)] \\
&=\mathbb{E}_S \mathbb{E}_L \mathbb{E}[Y \mid X, s, l] \\
&=\sum_s \sum_l E[Y \mid X, s, l] P(s) P(l).
\label{eq:counterfactual-1}
\end{aligned}
\end{equation}
where $s$/$l$ is a $S$/$L$-independent value. $d o$ interventions involve manipulating one or more variables to investigate causal relationships \cite{pearl2009causality}, where $d o\left(L:=l\right)$ signifies setting the value of variable $L$ to $l$ and observing the outcome. Note that, causal sufficiency is an essential assumption for Equation (\ref{eq:counterfactual-1}), \ie, the exogenous variables $U_i$ are jointly independent: $P(U_1, \ldots, U_n) = P(U_1) \times P(U_2) \times \cdots \times P(U_n)$. Thanks to the causal sufficiency assumption, the endogenous variables $\mathcal{V}$ in $\mathcal{M}$ can be formulated as a causal/disentangled factorization: 
\begin{equation}
P\left(V_1, V_2, \ldots, V_n\right)=\prod_{i=1}^n P\left(V_i \mid \textrm{pa}({V_i})\right),
\label{eq:disentangled-PA}
\end{equation}
where $\textrm{pa}({V_i})$ are the parents of $V_i$. In the SGG task, confounders can be, for instance, the observed confounders such as the semantic confusion confounder and the long-tailed distribution confounder, as well as unobserved ones caused by missing labeled relationships and mislabeled relationships. The latter has been discussed in much literature \cite{HML,BPLSA,NICE,DLFE,PKO}. We therefore do not expect and cannot model a causal sufficient SCM for the SGG task due to the unobserved confounders. In accordance with this, we assume that $\mathcal{M}_{\widetilde{v}}$ is causal-insufficient (see Assumption 1), and thus its endogenous variables can only be formulated as an entangled factorization: 
\begin{equation}
P\left(V_1,V_2, \ldots, V_n\right)=\prod_{i=1}^n P\left(V_i \mid V_{i+1}, \ldots, V_n\right).
\end{equation}

\noindent \textbf{Assumption 2} (Sparse Mechanism Shift (SMS) \cite{bengio2019meta,Bengio2021Toward}). \textit {Small distribution changes tend to manifest themselves sparsely or locally in the causal/disentangled factorization (see Equation (\ref{eq:disentangled-PA})), that is, they should usually not affect all factors simultaneously. }

Assumption 2 tells us that for a disentangled factorization, a sparse operation allows the learner to remove the confounders and even generalize to unseen distributions. However, unfortunately, $\mathcal{M}_{\widetilde{v}}$ is causal-insufficient since the SGG task inevitably contains the unobserved confounders, and it, therefore, cannot benefit from the SMS hypothesis. In response to this challenge, we decouple causal modeling into two stages to achieve the goal of intervening in the endogenous variables sparsely:
\begin{equation}
\begin{aligned}
\mathbb{E}&[Y \mid X, d o\left(S:= s\right), d o\left(L:=l\right)] \\
&=\mathbb{E}_X[\underbrace{\mathbb{E}[Y^{\prime} \mid X, d o(S:=s)]}_{\text{stage \ 1}}  + \underbrace{\mathbb{E}[Y \mid X, Y^{\prime}, d o(L:=l)]}_{\text{stage \ 2}}].
\end{aligned}
\end{equation}
where stage 1 exploits the inherent sparse property of similar relationships, even under the condition of causal-insufficient assumption, it can achieve sparse perturbations on variable $S$ to remove the semantic confusion confounder as well as learn a disentangled representation of $\mathcal{M}_{\widetilde{v}}$ at the same time, see Section \ref{sec3.2}. Based on the disentangled factorization obtained, we then, in stage 2, manipulate variable $L$ in a sparse way to remove the long-tailed distribution confounder, see Section \ref{sec3.3}. Both stages are sparse interventions, thereby satisfying the SMS assumption, which allows our method to achieve unbiased prediction while protecting the performance of head relationships. Specifically, stage 1, involving interventions on similar relationships, naturally doesn't harm head relationships. Moreover, the adjustment mechanism in stage 2, adaptively learned from stage 1, further ensures the protection of head relationships.

\subsection{Causal representation learning}
\label{sec3.2}
\subsubsection{Population Loss}
\label{sec3.2.1}
In the SGG task, similar relationships are those with only slightly different visual and semantic features. However, existing SGG models perform poorly in discriminating these similar relationships, for instance, easily mispredicting \textit {standing on} as \textit {walking on} or vice versa. This is not surprising, as distinguishing these similar relationships is even challenging for humans. Naturally, one may be curious and then imagine: Would the above error still occur if \textit {standing on} and \textit {walking on} are no longer similar? While this only happens in our imagined spaces, it can be formally calculated by counterfactual (see Definition 4) in the causal inference paradigm:
\begin{equation}
\begin{aligned}
P(y|& x,d o(S:= s)) \\
&=P(y|x,d o(S:= s_1))-P(y|x,d o(S:= s_0)),
\label{eq:counterfactual}
\end{aligned}
\end{equation}
where $(x,y)$ is a particular value of $(X,Y)$, $(X,Y) \sim \mathcal{D} $; $s_1$ and $s_0$ indicate that the relationship $y$ is similar or dissimilar to other relationships, respectively. In fact, the above counterfactual formulated in Equation (\ref{eq:counterfactual}) simulates the potential outcomes of different interventions, \ie, $d o(S:= s_1)$ and $d o(S:= s_0)$. It is critical because one can often benefit from imagining; for instance, Einstein's thought experiment brought the Special Theory of Relativity to the world. Despite being promising, however, calculating Equation (\ref{eq:counterfactual}) takes a lot of work. TDE \cite{TDE} is highly relevant to our work, which simulates two interventions by inferring pre/post-processed inputs to obtain counterfactual results. However, it requires two model inferences for each input, thereby introducing unbearable costs. In contrast, Average Treatment Effect (ATE) \cite{ATE} estimates the counterfactuals in one shot by leveraging statistical knowledge. Thanks to its high estimation efficiency, ATE is a commonly used technique in causal inference, such as exploring the ATE estimation with binary treatments and continuous outcomes in \cite{ATE_ADD_1} and discussing the propensity score if the average treatment effect is identifiable from observational data in \cite{ATE_ADD_2}. Inspired by ATE, in this paper, we use statistical knowledge from the observed data $\mathcal{D}$ to estimate counterfactuals:
\begin{equation}
\begin{aligned}
\mathbb{E} & [y |x,d o(S:= s)] \\
&=\mathbb{E}_X[\mathbb{E}(Y|X,d o(S:= s_1))-\mathbb{E}(Y|X,d o(S:= s_0))].
\label{eq:statis-method}
\end{aligned}
\end{equation}
 
\noindent \textbf{Definition 5} (Population in SGG).\textit { Let $y = \{y_1,y_2,\cdot\cdot\cdot,y_\emph{K} \}$ be relationship categories in observed data $\mathcal{D}$, and let $\textbf{P}_{\alpha}^{y_i}$ be population of $y_i$. Then, $\textbf{P}_{\alpha}^{y_i}$ is a relationship set containing the $\alpha$ most similar relationships to $y_i$. }

Formally, we first extract knowledge $\mathcal{P}_{\alpha}$, $\mathcal{P}_{\alpha} = \{ \textbf{P}_{\alpha}^{y_i} \}_{i=1}^{\emph{K}}$, from the observed data (the calculation of $\mathcal{P}_{\alpha}$ is placed in Section \ref{sec3.2.2}). $\textbf{P}_{\alpha}^{y_i}$ is the population of relationship $y_i$, a relationship set containing the statistical knowledge of similar relationships; see Definition 5. Inspired by penalized head categories in \cite{logitadjustment,add_longtail_2,add_longtail_3}, here we punish similar relationships based on knowledge $\mathcal{P}_{\alpha}$. Specifically, we discard the widely used cross-entropy loss $\ell$ and supervise the SGG model $f$ through the proposed Population Loss (P-Loss) $\hat{\ell}$:
\begin{equation}
\begin{aligned}
\hat{\ell}(\mathcal{P}_{\alpha}, y, f(x))=\log [1&+ \sum_{y^{\prime} \in \textbf{P}_{\alpha}^y}\frac{\pi_{y^{\prime}}}{\pi_y} \times e^{(f_{y^{\prime}}(x)-f_y(x))} \\
& +\sum_{y^{\prime} \notin \textbf{P}_{\alpha}^y , y^{\prime} \neq y} e^{(f_{y^{\prime}}(x)-f_y(x))}],
\label{eq:PLloss}
\end{aligned}
\end{equation}
\begin{equation}
\theta^*=\underset{\theta}{\arg \min } \mathbb{E}[\underset{(x, y) \sim \mathcal{D}}{\mathbb{E}} \hat{\ell}(\mathcal{P}_{\alpha}, y, f(x))],
\end{equation}

where $\pi$ is category frequencies on the observed data $\mathcal{D}$ and $\theta^*$ is the parameter used to parameterize SGG model $f_{\theta^*}$. $x$ and $y$ ($y = \{y_i\}_{i=1}^{\emph{K}}$) are the input (\eg, image) and output relationship categories, respectively. As an example, for relationship $y_i$, the P-Loss $\hat{\ell}$ penalizes its confusing relationships with the help of statistical knowledge $\pi$ and $\textbf{P}_{\alpha}^{y_i}$ extracted from the observed data $\mathcal{D}$. The penalty term in Equation (\ref{eq:PLloss}) can be seen as $d o(S:= s)$ in Equation (\ref{eq:counterfactual}) since it intervenes in the sparse $\mathcal{P}_{\alpha}$ and makes the model more capable of distinguishing between similar relationships. In other words, P-Loss can remove the confounder $S$ in $\mathcal{M}_{\widetilde{v}}$. Thus, the counterfactual can be estimated by the statistical knowledge $\mathcal{P}_{\alpha}$ as: 
\begin{equation}
\begin{aligned}
P(y| x,d o(S:= s)) &= P(y| x,\mathcal{P}_{\alpha},\pi) \\
&= f_{\theta^*}\left(x\right) .
\end{aligned}
\end{equation}

\noindent \textbf{Assumption 3} (Similar relationships are sparse). \textit { Let $y = \{y_i\}_{i=1}^{\emph{K}}$ be relationships in observed data $\mathcal{D}$. For any relationship $y_i$ ($i \in\{ 1,2, \cdots, \emph{K} \}$), there exist $k$ relationships similar to it. Then, it holds that $k \ll \emph{K}$.}

Despite achieving the goal of calculating the counterfactuals, however, it is critical to note that our causal-insufficient assumption determines that our manipulation ($d o(S:= s)$) in  Equation (\ref{eq:statis-method}) may perturb other variables simultaneously since the entangled factorization of $\mathcal{M}_{\widetilde{v}}$ does not satisfy the SMS hypothesis. Fortunately, in this paper, we empirically find that similar relationships hold the sparse property (see Assumption 3). Based on our observations, relationships within the SGG dataset are often similar to a few specific relationships but not to most others. For example, \textit {standing on} is only similar to \textit {on}, \textit {walking on}, \etc., but differs from most other relationships. Therefore, Assumption 3, which shows similar relationships have the sparse property that SMS highlighted, guarantees that even an entangled factorization can be intervened sparsely on the confounder $S$. In other words, even if $\mathcal{M}_{\widetilde{v}}$ is causal-insufficient, our proposed P-Loss can still intervene in $S$ without worrying about perturbing other exogenous variables, such as confounder $L$. 

Furthermore, we argue that $d o(S:= s)$ partially disentangles the $\mathcal{M}_{\widetilde{v}}$ as it removes the confounder $S$ and allows us to get a better causal representation. Thus, the endogenous variables in the induced submodel $\mathcal{M}_{\widetilde{v}}^{{S}}$ can be roughly formulated as a disentangled factorization:
\begin{equation}
P(X,Y,L)\doteq P(X) \times P(Y) \times P(L).
\label{eq:disentangled-XYL}
\end{equation}

Disentangled factorization is considered to be the key to representation learning due to its potential in abstract reasoning, interpretability, generalization to unseen scenarios, \etc. Although it has attracted significant attention, evaluating the disentangled representation is still challenging \cite{disentangled}. We will design experiments in the ablation study (Section \ref{Ablation}) to demonstrate our disentangled claim in Equation (\ref{eq:disentangled-XYL}).

\subsubsection{Calculate the relationship-populations}
\label{sec3.2.2} 
As a supplement to Section \ref{sec3.2.1}, this section shows how to calculate the relationship-populations $\mathcal{P}_{\alpha}$. For $\mathcal{P}_{\alpha}$, we give three assumptions (Assumptions 4-6) based on the inspirations of causality as well as the hallmarks of the relationships in the SGG task.

\noindent \textbf{Assumption 4} (Relationship-population is learner independent). \textit { Let $f_{\theta_1}$ and $f_{\theta_2}$ be SGG models parameterized by $\theta_1$ and $\theta_2$, respectively. Then, $\mathbb{E}[\mathcal{P}_{\alpha}^{'} \mid f_{\theta_1}]=\mathbb{E}[\mathcal{P}_{\alpha}^{''} \mid f_{\theta_2}]$. }

\noindent \textbf{Assumption 5} (Relationship-population is distribution insensitive). \textit { Let $D_{obs}^1$ and $D_{obs}^2$ be two observed datasets. Then, $\mathbb{E}[\mathcal{P}_{\alpha}^{'} \mid D_{obs}^1]=\mathbb{E}[\mathcal{P}_{\alpha}^{''} \mid D_{obs}^2]$. }

\noindent \textbf{Assumption 6} (Relationship-population is asymmetric). \textit { Let $y_i$ and $y_j$ be two relationships. Then, $y_i \in \textbf{P}_{\alpha}^{y_j} \nLeftrightarrow y_j \in \textbf{P}_{\alpha}^{y_i}$. }

Assumption 4 states that whether two relationships are similar is irrelevant to the SGG model. \textit {Standing on} and \textit {walking on}, for instance, should share the same features no matter what model we use. In light of this, we should not use any SGG model when calculating $\mathcal{P}_{\alpha}$. Assumption 5 illustrates no correlation between the distribution of two relationships and their similarity. We highlight this because we observe that the SGG dataset often suffers from the long-tailed distribution issue at both the relationship and object levels, which may perturb the calculation procedure of $\mathcal{P}_{\alpha}$. Assumption 6 is inspired by the fact that cause and effect are directed, \ie, the cause can determine the effect, but not vice versa.

To satisfy Assumption 4, we design a model-agnostic relationship feature extraction method. Consider two objects, $o_i$ and $o_j$, whose bounding boxes are $\left[b_{\bar{x}}^i, b_{\bar{y}}^i, b_{h}^i, b_{w}^i\right]$ and $[b_{\bar{x}}^j, b_{\bar{y}}^j, b_{h}^j, b_{w}^j]$, respectively. Where, as an example, for the bounding box of $o_i$, $(b_{\bar{x}}^i, b_{\bar{y}}^i)$ is the center point, and $b_{w}^i$ and $b_{h}^i$ are the width and height. We denote the model-agnostic feature of the relationship between these two objects as $\psi_{<o_i,o_j>}$:
\begin{equation}
\begin{aligned}
&{[\frac{2(b_{\bar{x}}^i+b_{\bar{x}}^j)-(b_{w}^i+b_{w}^j)}{4 b_{h}^i}, \frac{2(b_{\bar{y}}^i+b_{\bar{y}}^j)-(b_{h}^i+b_{h}^j)}{4 b_{h}^i}} , \\
& \ \ \frac{2(b_{\bar{x}}^i+b_{\bar{x}}^j)+(b_{w}^i+b_{w}^j)}{4 b_{h}^i}, \frac{2(b_{\bar{y}}^i+b_{\bar{y}}^j)+(b_{h}^i+b_{h}^j)}{4 b_{h}^i} , \frac{b_{h}^j}{b_{h}^i}, \frac{b_{w}^j}{b_{w}^i}].
\label{eq:feature}
\end{aligned}
\end{equation}
Our proposed model-agnostic feature emphasizes the relative position between object pairs, which is inspired by the fact that it is intrinsically linked to the relationships in SGG. For example, the object pairs of \textit {standing on} are up-down, while \textit {behind} is front-back. Thanks to the molecules of Equation (\ref{eq:feature}), $\psi_{<o_i,o_j>}$ is position-insensitive, as the upper left corners of all object pairs are moved to the same coordinate. Besides, the denominator of Equation (\ref{eq:feature}) normalizes the object pairs, ensuring that $\psi_{<o_i,o_j>}$ is scale-insensitive. The position/scale-insensitive design in our model-agnostic feature extraction method can overcome the problem that the distance of the lens can make the same relationship vary greatly, thereby generalizing to unseen object pairs.

Before extracting the relationship features in the observed data using the above method, however, there is a problem that needs to address: The object-level long-tailed distribution problem may perturb the model-agnostic feature extraction. Consider an example with $90\%$ \textit {$<$people, standing on, road$>$} and $10\%$ \textit {$<$people, standing on, beach$>$} in the observed data. It is fusing the features of \textit {standing on} will undoubtedly bias towards \textit {$<$people, standing on, road$>$} due to its dominance in the observed data, which is detrimental to extracting the feature of \textit {standing on}. We address this problem by extracting the object-to-object level features $\boldsymbol{\xi}_y$ and then normalizing them. Our method is inspired by Inverse Probability Weighting (IPW) \cite{IPW}, a bias correction method commonly used in statistics, econometrics, epidemiology, \etc. As a result of this improvement, the proposed method satisfies Assumption 5 since it eliminates the disturbance of distribution from the feature extraction. Specifically, $\boldsymbol{\xi}_y \in \mathbb{R}^4$ $(\emph{C} \times \emph{C} \times \emph{K} \times 6)$ is a four-dimensional statistic:
\begin{equation}
\boldsymbol{\xi}_y=\left[\begin{array}{cccc}
\boldsymbol{\xi}_y^{(1,1)} & \boldsymbol{\xi}_y^{(1,2)}  & \cdots & \boldsymbol{\xi}_y^{(1, \emph{C})} \\
\cdots & \cdots & \cdots & \cdots\\
\boldsymbol{\xi}_y^{(\emph{C}, 1)} & \boldsymbol{\xi}_y^{(\emph{C}, 2)}  & \cdots & \boldsymbol{\xi}_y^{(\emph{C}, \emph{C})}
\end{array}\right],
\end{equation}
where $\boldsymbol{\xi}_y^{(i, j)} = \{ \xi_{y_t}^{(i, j)} \}_{t=1}^{\emph{K}}$ is the normalized features of relationship ${<o_i, y_t, o_j>}$, and it can be calculated as:
\begin{equation}
\xi_{y_t}^{(i, j)}= \xi_{y_t}^{<o_i, o_j>} / |\xi_{y_t}^{<o_i, o_j>}|,
\end{equation}
$\xi_{y_t}^{<o_i, o_j>}$ and $|\xi_{y_t}^{<o_i, o_j>}|$ are the fusion features and numbers of all relationships $<o_i, y_t, o_j>$ in the observed data $\mathcal{D}$, respectively. We then calculate the feature of each relationship, for instance, for the $t$-th relationship $\xi_{y_t}$:
\renewcommand{\thefootnote}{\fnsymbol{footnote}}
\begin{equation}
\xi_{y_t}=\sum_{i=1}^\emph{C} \sum_{j=1}^\emph{C} \xi_{y_t}^{(i, j)} / \emph{C}^2 \footnote[4]{ Object pair $<o_i,o_j>$ theoretically has $\emph{C}^2$ triplet categories. However, the annotations in SGG dataset are very sparse, \ie, $<o_i,o_j>$ usually only covers a few triplet categories, resulting in a much smaller number of triplet categories than $\emph{C}^2$. As a result, the numerator term in Equation (\ref{eq:RelationFeature}) should be determined by the observed data $\mathcal{D}$. For instance, for $y_t$, the numerator term should be the number of triplet categories composed of $y_t$ in $\mathcal{D}$.}.
\label{eq:RelationFeature}
\end{equation}

For relationship-populations $\mathcal{P}_{\alpha}$, $\mathcal{P}_{\alpha} = \{ \textbf{P}_{\alpha}^{y_i} \}_{i=1}^{\emph{K}}$, the population of $y_t$ can be calculated as:
\begin{equation}
\textbf{P}_{\alpha}^{y_t}=\underset{t, t^{\prime} \in \{ 1,2, \cdots, \emph{K} \}, t \neq t^{\prime}}{\arg \operatorname{smal} \alpha}\left\|\xi_{y_t}-\xi_{y_{t^{\prime}}}\right\|,
\label{eq:kernel}
\end{equation}
where $\arg \operatorname{smal} \alpha$ is a computation kernel that selects similar relationships based on feature distances. As an example, Equation (\ref{eq:kernel}) takes the $\alpha$ relationships with the smallest feature distance from $\xi_{y_t}$. Our method guarantees that the head and tail relationship categories have the same population scale and thus satisfy Assumption 6. As such, different relationships yield different feature distance thresholds in Equation (\ref{eq:kernel}), resulting in asymmetric relationship-populations.

\subsection{Causal calibration learning}
\label{sec3.3}

\subsubsection{Adaptive Logit Adjustment}
\label{sec3.3.1}
Fig.~\ref{motivation} illustrates the severe long-tailed distribution problem in the SGG task. The current SGG models, therefore, easily predict informative fine-grained relationships as less informative coarse-grained relationships. For instance,  \textit {looking at} is predicted as  \textit {near}. To end this, let us seek the imagination again: If one collected the balanced data, or, in particular,  \textit {looking at} and  \textit {near} share the same distribution in the observed data $\mathcal{D}$, will the above error still occur? Similar to Equation (\ref{eq:counterfactual}), this question can also be answered with the counterfactual: 
\begin{equation}
\begin{aligned}
P(y|& x,d o(L:= l)) \\
&=P(y|x,d o(L:= l_1))-P(y|x,d o(L:= l_0)),
\label{eq:counterfactual2}
\end{aligned}
\end{equation}
where $l_1$ and $l_0$ represent the head and tail categories, respectively; as such, Equation (\ref{eq:counterfactual2}) simulates the potential outcomes of different interventions, \ie, $d o(L:= l_1)$ and $d o(L:= l_0)$. Inspired by logit adjustment \cite{post_hoc1,post_hoc2,post_hoc3}, in which the class prior knowledge (also known as adjustment factors) extracted from the training data are used to adjust the model results, we extract the statistical knowledge from the observed data $\mathcal{D}$ via model $f_{\theta^*}$ to estimate counterfactuals:
\begin{equation}
\begin{aligned}
\mathbb{E} & [y |x,d o(L:= l)] \\
&=\mathbb{E}_X[\mathbb{E}(Y|X,d o(L:= l_1))-\mathbb{E}(Y|X,d o(L:= l_0))].
\label{eq:stage2counterfactual}
\end{aligned}
\end{equation}

Specifically, we leverage the extracted statistical knowledge, adjustment factors $\textbf{T}_{\beta}$, to maximize the recall rate of $f_{\theta^*}$ on the observed data (the computation of $\textbf{T}_{\beta}$ is placed in Section \ref{sec3.3.2}). Compared to existing logit adjustment methods, our adjustment factors $\textbf{T}_{\beta}$ not only extract knowledge from $\mathcal{D}$ but, more importantly, it fits adaptively to the SGG model $f_{\theta^*}$. Holding this advantage, our adjustment factors $\textbf{T}_{\beta}$ outperform the traditional adjustment method by a clear margin (see the experiments in Section \ref{Ablation}). Despite this, the knowledge extracted directly from $f_{\theta^*}$ and $\mathcal{D}$ is still suboptimal. We think this is because background relationships will perturb the model training, resulting in 1) the logits of the foreground relationships being less discriminative; 2) the logits of alternating positive and negative make it impossible for the learned factors to adjust to some predictions correctly (see the qualitative results in Fig.~\ref{AblationResult}). Where foreground relationships in the SGG task are those within annotated triplets in the observed data, and background relationships are the ones that are absent between object pairs. To overcome these problems, we augment the logits of $f_{\theta^*}$:
\begin{equation}
\tilde{f}_{\theta^*, y}(x)=e^{f_{\theta^*, y}(x)} \times f_{\theta^*, y}^{\textrm{bg}}(x),
\label{eq:logit-agument}
\end{equation}
where $f_{\theta^*, y}^{\textrm{bg}}(x)$ is the logit of the corresponding background relationship output by $f_{\theta^*}$. $f_{\theta^*, y}^{\textrm{bg}}(x)$ acts as a guidance term that can make the augmented logits $\tilde{f}_{\theta^*, y}(x)$ more discriminative, which is inspired by the impressive performance of the traditional adjustment methods in the simple classification task. Augmented logits allow us to get better adjustment factors, and then the final prediction $y_x$ of input $x$ can be calculated as:
\begin{equation}
y_x=\underset{y \in\{ y_1, \cdots, y_\emph{K} \}}{\arg \max }\{(\tilde{f}_{\theta^*, y}(x) \times \textbf{T}_{\beta})_{y \in \textbf{T}_{\beta}} \cap(\tilde{f}_{\theta^*, y}(x))_{y \notin \textbf{T}_{\beta}}\}.
\label{eq:adjust}
\end{equation}

Consider a typical false prediction: For an input $x$ belonging to the tail category $y_i$, the largest and next largest output logits correspond to $y_j$ and $y_i$, where $y_j$ is a head category. However, our proposed adjustment factors can correct this false prediction by penalizing the logits corresponding to the head categories and encouraging the tail categories, thus, eliminating the negative effect caused by the long-tailed distribution problem. Our proposed AL-Adjustment acts as $d o(L:= l)$ in Equation (\ref{eq:stage2counterfactual}) to remove the confounder $L$ in the induced submodel $\mathcal{M}_{\widetilde{v}}^{{S}}$. Therefore, the estimated counterfactual by the statistical knowledge $\textbf{T}_{\beta}$ is: 
\begin{equation}
\begin{aligned}
P(y| x,d o(L:= l)) &= P(y| x,\textbf{T}_{\beta},\tilde{f}_{\theta^*}) \\
&=\tilde{f}_{\theta^*, y}(x) \times \textbf{T}_{\beta}
\end{aligned}.
\end{equation}

In Section \ref{sec3.2.1}, we show that $\mathcal{M}_{\widetilde{v}}^{{S}}$ can be decomposed into a disentangled factorization. As a result, manipulating the factor in $\mathcal{M}_{\widetilde{v}}^{{S}}$, in most cases, should not affect all factors simultaneously (SMS hypothesis, see Assumption 2). We therefore argue that $d o(L:= l)$ in Equation (\ref{eq:stage2counterfactual}) does not affect the exogenous variables $X$ and $Y$, and then the induced submodel $\mathcal{M}_{\widetilde{v}}^{{S},{L}}$ obtained in this stage can be further roughly formulated as a disentangled factorization:
 \begin{equation}
P(X,Y)\doteq P(X) \times P(Y).
\label{eq:disentangled-XY}
\end{equation}

We will design experiments in the ablation study (Section \ref{Ablation}) to demonstrate our disentangled claim in Equation (\ref{eq:disentangled-XY}).

\subsubsection{Calculate the adjustment factors}
\label{sec3.3.2}
This subsection shows how to extract statistical knowledge $\textbf{T}_{\beta}$ from the observed data $\mathcal{D}$ and the SGG model $\tilde{f}_{\theta^*}$, which can be used to adjust the logits to remove the confounder $L$ in submodel $\mathcal{M}_{\widetilde{v}}^{{S}}$. For adjustment factors $\textbf{T}_{\beta}$, we have two assumptions (Assumptions 7-8).  

\noindent \textbf{Assumption 7} (Adjustment effect should be sparse). \textit { Let $\textbf{T}_{\beta}^{y_i}$ and $\textbf{T}_{\beta}^{y_j}$ are adjustment factors of $i$-th and $j$-th prediction logits, respectively. Then, $P(y_i \mid x)=P(y_i \mid x, \textbf{T}_{\beta}^{y_j})$, $P(y_j \mid x)=P(y_j \mid x, \textbf{T}_{\beta}^{y_i})$. }

\noindent \textbf{Assumption 8} (Adjustment factors should be independent of each other). \textit { Let $\textbf{T}_{\beta}^{y_i}$ and $\textbf{T}_{\beta}^{y_j}$ are adjustment factors of $i$-th and $j$-th prediction logits, respectively. Then, $P(y \mid x,Max(\textbf{T}_{\beta}^{y_i},\textbf{T}_{\beta}^{y_j})) = Max(P(y \mid x, \textbf{T}_{\beta}^{y_i}),P(y \mid x, \textbf{T}_{\beta}^{y_j}))$, where $Max(\cdot | \cdot)$ is a computation kernel to take the maximum value of the corresponding positions of the two sets.}

Assumption 7 is inspired by the SMS hypothesis, and it holds due to the disentangled factorization (Equation (\ref{eq:disentangled-XYL})) obtained in stage 1. This assumption also stems from our insight that the false predictions in most cases belong to the largest few logits (see Table~\ref{tab:logitordering}). As such, Assumption 7 allows us to correct the false predictions with sparse adjustment factors. However, the existing methods adjust all logits. Assumption 8 views the SMS hypothesis through the relationship level to highlight the causality between the relationships. Here is an intuition for this assumption: To correct any false prediction logits of a binary classification task, we only have to adjust one of these two logits. Assumption 8 allows us to learn the adjustment factors of each relationship independently.

Our proposed adaptive adjustment factors $\textbf{T}_{\beta}$ is a two-dimensional $(\emph{K} \times \beta)$ matrix:
\begin{equation}
\textbf{T}_{\beta}=\left[\begin{array}{cccc}
\emph{T}_{\beta}^{y_1,l_1} & \emph{T}_{\beta}^{y_1,l_2} & \cdots &\emph{T}_{\beta}^{y_1,l_\beta} \\
\cdots & \cdots & \cdots & \cdots \\
\emph{T}_{\beta}^{y_\emph{K},l_1} & \emph{T}_{\beta}^{y_\emph{K},l_2} & \cdots &\emph{T}_{\beta}^{y_\emph{K},l_\beta} 
\end{array}\right],
\label{eq:adjust_matrix}
\end{equation}
where $\emph{T}_{\beta}^{y_i,l_j}$ adjusts the $j$-th largest prediction logit to correspond to the $i$-th relationship, and it can be calculated as:
\begin{equation}
\begin{aligned}
\emph{T}_{\beta}^{y_i,l_j}=\underset{\emph{T}_{\beta}^{y_i,l_j} \in \emph{T}}{\arg \max } ( &\underbrace{{\underset{(x, y) \sim \mathcal{D}^{y_i,l_j}}{\mathds{TP}}(\tilde{f}_{\theta^*, y}(x) \times \emph{T}_{\beta}^{y_i,l_j})}}_{\text{true \ prediction \ with \ adjustment}}- \\
&\underbrace{{\underset{(x, y) \sim  \mathcal{D}^{y_i,l_j}}{\mathds{TP}}(\tilde{f}_{\theta^*, y}(x)) \ ) }}_{\text{true \ prediction \ without \ adjustment}},
\label{eq:adjust_tp}
\end{aligned}
\end{equation}
where $\emph{T} \in \mathbb{R}$ and ${\underset{(X, Y)}{\mathds{TP}}(f)}$ is a computation kernel that calculates the true prediction numbers (\eg, recall rate (R@K) in SGG task) of model $f$ on dataset $(X,Y)$. $\mathcal{D}^{y_i,l_j}$ is all relationships with the $j$-th largest prediction logit to correspond to the $i$-th category reasoned by $f_{\theta^*}$, and it can be further divided into true predictions $\mathcal{TD}_{\textrm{obs}}^{y_i,l_j}$ and false predictions $\mathcal{FD}_{\textrm{obs}}^{y_i,l_j}$. Thus, our method is to maximize the recall rate of model $f_{\theta^*}$ on the observed data $\mathcal{D}$ by the adjustment factors learned in Equation (\ref{eq:adjust_tp}).

\begin{figure}
	\footnotesize\centering
	\centerline{\includegraphics[width=1\linewidth]{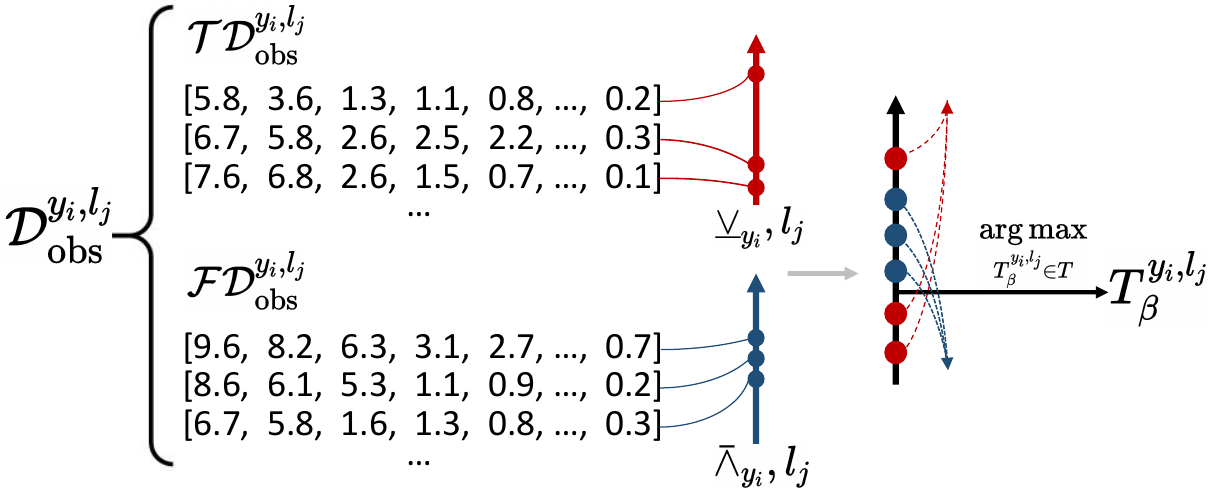}}
    \vspace{-0.2cm}
    \caption{The proposed upper-lower bound-based method of calculating Equation (\ref{eq:adjust_tp}). Each False/True prediction logit corresponds to an upper/lower bound. The most optimal adjustment factor needs to satisfy the most bounds.}
    \label{up-low-bounds}
    \vspace{-0.3cm}
\end{figure}

We then propose an upper-lower bound-based method to compute Equation (\ref{eq:adjust_tp}) quickly. As shown in Fig.~\ref{up-low-bounds}, for each relationship in $\mathcal{TD}_{\textrm{obs}}^{y_i,l_j}$, we can compute a lower bound that keeps the correct prediction. Similarly, we can obtain for each relationship in $\mathcal{FD}_{\textrm{obs}}^{y_i,l_j}$ an upper bound that can adjust it to the correct prediction. We denote the lower and upper bounds of $\mathcal{D}^{y_i,l_j}$ as $\veebar_{y_i,l_j}$ and $\barwedge_{y_i,l_j}$, respectively. It is clear that we only need to let $\emph{T}_{\beta}^{y_i,l_j}$ satisfy the most bounds to maximize the number of correct predictions. Therefore, maximizing the recall rate of model $f_{\theta^*}$ on the observed data $\mathcal{D}$ by the adjust factor is equivalent to another task that finds the factor that satisfies the most bounds in $\veebar_{y_i,l_j}$ and $\barwedge_{y_i,l_j}$. As a result, $\emph{T}_{\beta}^{y_i,l_j}$ in Equation (\ref{eq:adjust_tp}) can also be calculated by:
\begin{equation}
\begin{aligned}
\emph{T}_{\beta}^{y_i,l_j}=\underset{t \in \emph{T}}{\arg \max } ( \sum_{m=1}^{|\veebar_{y_i,l_j}|} \mathds{1}(t \geq \veebar_{y_i,l_j}^m)+\\
\sum_{n=1}^{|\barwedge_{y_i,l_j}|} \mathds{1}(t < \barwedge_{y_i,l_j}^n)),
\label{eq:adjust_factors}
\end{aligned}
\end{equation}
where $\mathds{1}( \cdot )$ is an indicator function (equals 1 if the expression is \textit{true} and 0 for \textit{false}) and $| \cdot |$ is the length/size of the given set/list. However, we further find that the long-tailed distribution problem may perturb the adjustment effect of $\emph{T}_{\beta}^{y_i,l_j}$. Specifically, if $y_i$ is a head category, $|\veebar_{y_i,l_j}| \ll |\barwedge_{y_i,l_j}|$, and if it is a tail category, then $|\barwedge_{y_i,l_j}| \gg |\veebar_{y_i,l_j}|$. It is due to biased training caused by the skewed distribution. To this issue, for relationship $y_i$, we randomly sample the same number (\eg, $min(|\veebar_{y_i,l_j}|, |\barwedge_{y_i,l_j}|)$) of bounds $\veebar_{y_i,l_j}^{'}$ and $\barwedge_{y_i,l_j}^{'}$ separately from the original lower/upper bounds to ensure unbiased adjustment factors. Therefore, Equation (\ref{eq:adjust_factors}) will be adjusted to:
\begin{equation}
\begin{aligned}
\emph{T}_{\beta}^{y_i,l_j}=\underset{t \in \emph{T}}{\arg \max } ( \sum_{m=1}^{min(|\veebar_{y_i,l_j}| , |\barwedge_{y_i,l_j}|)} \mathds{1}(t \geq \veebar_{y_i,l_j}^{' m})+\\
\sum_{n=1}^{min(|\veebar_{y_i,l_j}| , |\barwedge_{y_i,l_j}|)} \mathds{1}(t < \barwedge_{y_i,l_j}^{' n})).
\label{eq:adjust_factors_unbias}
\end{aligned}
\end{equation}

Note that in Equation (\ref{eq:adjust_factors_unbias}), $\emph{T}_{\beta}^{y_i,l_j}$ is an interval with extremely close upper and lower bounds, thereby selecting any value within this interval as the adjustment factor has a negligible impact on the results. Consequently, in this paper, we randomly sample a value from $\emph{T}_{\beta}^{y_i,l_j}$ as the learned adjustment factor. Finally, for each relationship, we learn only $\beta$ adjustment factors corresponding to the 1-$\beta$ positions in the prediction logits. Thus, this sparse adjustment mechanism enables our method to satisfy Assumption 7. Meanwhile, the adjustment factors for each relationship are independently learned by Equation (\ref{eq:adjust_factors_unbias}), so our method satisfies Assumption 8.

\subsection{Discussion}
\label{sec3.4}
This subsection first shows that our method is Fisher consistent, \ie, models based on popular learning strategies (\eg, empirical risk minimization (ERM)) lead to the Bayes optimal rule of classification that minimizes the balanced error \cite{Fisher1,Fisher2}. This is very important for the SGG task, as it prevents the model from heading down a confusing path, \ie, biased towards predicting head categories for a high recall rate. We then highlight the differences and advantages of the proposed causal framework with existing methods.

\subsubsection{Fisher  consistency}
To demonstrate that our method is Fisher consistent, we start with the Bayes perspective. \cite{logitadjustment} thoroughly explored the relationship between the posterior probability of the balanced class-probability function $P^{\mathrm{bal}}(y \mid x)$ and the unbalanced one $P(y \mid x)$, and it defined $P^{\mathrm{bal}}(x \mid y) \propto P(x \mid y) / P(x)$. In the SGG task, however, we find that the models suffer from confounders other than the long-tailed distribution problem, such as semantic confusion confounder, as well as unobserved ones like missing labeled relationship confounder and mislabeled relationship confounder. As such, here we define: 
\begin{equation}    
P^{\mathrm{bal}}(x \mid y) \propto P(x \mid y) / P(x)P(S)P(U_o),
\end{equation}
where $U_o$ is unobserved confounders. Also, consider:
\begin{equation}
P^{\mathrm{bal}}(y \mid x)=(P^{\mathrm{bal}}(x \mid y)  P^{\mathrm{bal}}(y)) / P^{\mathrm{bal}}(x) , 
\end{equation}
we therefore have:
\begin{equation}
\begin{aligned}
P^{\mathrm{bal}}(y \mid x) &\propto (P(x \mid y)  P(X)  P^{\mathrm{bal}}(y)) / (P(X) \\
&  \ \ \ \ \ \  P(Y)  P(S)  P(U_o)  P^{\mathrm{bal}}(x)).
\end{aligned}
\end{equation}

For fixed class-conditionals $P(x | y)$, the optimal predictions will not be affected by $P(Y)$ \cite{logitadjustment}, hence:
\begin{equation}
P^{\mathrm{bal}}(y \mid x) \propto P(y \mid x)  / ( P(S)  P(U_o)  P^{\mathrm{bal}}(x)).
\label{eq:Fisher_left}
\end{equation}
Then, according to the SMS hypothesis (Assumption 2) and small distribution changes hypothesis in \cite{Bengio2021Toward}, there exists an intervention $\mathcal{I}$ such that: 
\begin{equation}
\underset{y \in\{ y_1, \cdots, y_\emph{K} \}}{\arg \max } f_{\theta^{\mathcal{I}}}(x) =\underset{y \in\{ y_1, \cdots, y_\emph{K} \}}{\arg \max }(\tilde{f}_{\theta^*, y}(x) \times \textbf{T}_{\beta}). 
\end{equation}

Note that we cannot model the intervention $\mathcal{I}$ directly since the induced submodel $\mathcal{M}_{\widetilde{v}}$ can only be formulated as an entangled factorization. We define the adjustment factors corresponding to intervention $\mathcal{I}$ as $\textbf{T}_{\beta}^\mathcal{I}$, that is:
\begin{equation}
\underset{y \in\{ y_1, \cdots, y_\emph{K} \}}{\arg \max } f_{\theta^{\mathcal{I}}}(x) =\underset{y \in\{ y_1, \cdots, y_\emph{K} \}}{\arg \max }(\tilde{f}_{\theta^*, y}(x) \times \textbf{T}_{\beta}^\mathcal{I}).
\end{equation}
Based on the  Theorem 1 in the \cite{Theorem1}, we have

$\operatorname{argmax}_{y \in\{ y_1, \cdots, y_\emph{K} \}} \tilde{f}_{\theta^*, y}(x) = \operatorname{argmax}_{y \in\{ y_1, \cdots, y_\emph{K} \}} P (x \mid y)$, 
thus:  
\begin{equation}
\underset{y \in\{ y_1, \cdots, y_\emph{K} \}}{\arg \max } f_{\theta^{\mathcal{I}}}(x) =\underset{y \in\{ y_1, \cdots, y_\emph{K} \}}{\arg \max }((P(y \mid x)P(y)/P(x)) \times \textbf{T}_{\beta}^\mathcal{I}).
\label{eq:Fisher_right}
\end{equation}
Considering both Equation (\ref{eq:Fisher_left}) and Equation (\ref{eq:Fisher_right}), when 
\begin{equation}
\textbf{T}_{\beta}^\mathcal{I} \propto P(Y) / (P(S)P(X)P(U_o) P^{\mathrm{bal}}(x)) , 
\end{equation}
then 
\begin{equation}
\underset{y \in\{ y_1, \cdots, y_\emph{K} \}}{\arg \max } f_{\theta^{\mathcal{I}}}(x) = \underset{y \in\{ y_1, \cdots, y_\emph{K} \}}{\arg \max }P^{\mathrm{bal}}(y \mid x). 
\end{equation}
This means that our manipulations on confounder $S$ and confounder $L$ can lead to a minimal balanced error (mR@K in SGG task), and thus our method is Fisher consistent.

\subsubsection{Sparsity and independency}
Causal representation learning (stage 1) in our proposed causal framework is inspired by the loss-weighting methods \cite{Lossweighting1,Lossweighting2,Lossweighting3,li2022targeted}, and causal calibration learning (stage 2) is inspired by post-hoc adjustment approaches \cite{post_hoc1,post_hoc2,post_hoc3}. In all of these heuristic works, statistical priors (\eg, category frequencies) are extracted from the observed data to calibrate the decision boundaries. However, we leverage the extracted statistical knowledge to estimate the counterfactual to eliminate confounders $S$ and $L$. Where the statistical knowledge in stage 1 is extracted via the proposed model-agnostic method, and that of in stage 2 is adaptively extracted from the learned model $f_{\theta^*}$ and the observed data $\mathcal{D}$. Besides, our method differs fundamentally from these works in that the interventions using knowledge are sparse and independent, which is the key to preserving head category performance while pursuing the prediction of high-informative tail relationships.

\begin{figure}
	\footnotesize\centering
	\centerline{\includegraphics[width=1\linewidth]{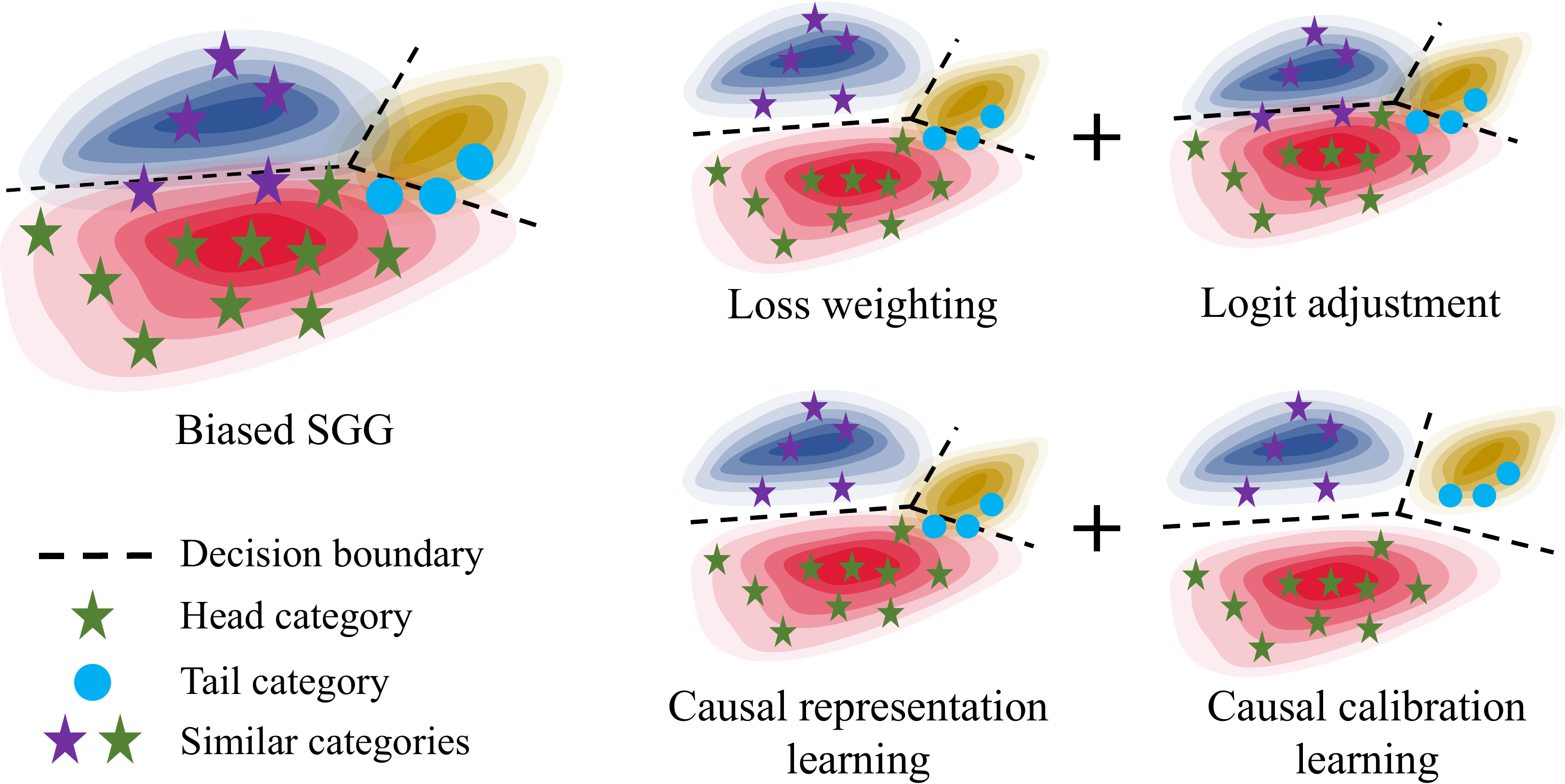}}
        \vspace{-0.2cm}
        \caption{The merging effect of statistical-based methods (top) and our proposed causal components (bottom).}
	\label{methodcomparison}
        \vspace{-0.3cm}
\end{figure}
Causal inference models the observed data with modular knowledge, and interventions on partial knowledge can achieve rapid distribution changes \cite{bengio2019meta}. These sparse perturbations simulate human learning, \ie, the reuse of most knowledge, and thus, have great potential for practical applications, especially for open-world learning. Both stage 1 and stage 2 of our causal framework are sparse, specifically, $\alpha$ in $\mathcal{P}_{\alpha}$ and $\beta$ in $\textbf{T}_{\beta}$. The former means that each relationship only takes the $\alpha$ most similar ones as its population. Therefore, Equation (\ref{eq:PLloss}) only sparsely adjusts the loss for very few relationships. The latter represents that only the top-$\beta$ predict logits will be adjusted. Thus, Equation (\ref{eq:adjust}) is a sparse adjustment technique.

Independent Causal Mechanisms (ICM) \cite{ICM} tells us that changing one causal mechanism does not change others. Note that ICM requires causal sufficiency, but the SGG task does not satisfy this. However, as analyzed in Section \ref{sec3.2.1}, our proposed P-Loss can intervene in confounder $S$ without losing the independent property due to the sparse nature of similar relationships. The result of stage 1 can be roughly formulated as a disentangled factorization. Furthermore, the different logit positions of $\textbf{T}_{\beta}$ are independently learned. These enable independent intervention in stage 2. 

In addition, \cite{logitadjustment} shows that loss-reweight and logit-reweight are identical, and merging them brings no further gain. The latter even cancels out the improvement from the former in some cases. However, the post-hoc adjustment factors in our causal framework are adaptively learned from the model obtained in the previous stage and thus always yield positive adjustment effects. More importantly, our method can make the decision boundaries between similar relationships clearer, which traditional methods cannot achieve. We show the two above merge routes in Fig.~\ref{methodcomparison} to compare the boundary adjustment processes. 

\begin{table*}[htbp]
  \centering
  \caption{Quantitative results (mR@K) of our method and other baselines on the MotifsNet backbone}
  \begin{threeparttable}
  \vspace{-0.4cm}
  \setlength\tabcolsep{2pt}
\setlength\arrayrulewidth{0.2mm}
    \resizebox{17cm}{!}{\begin{tabular}{l|l|cccc|cccc|cccc}
    \toprule
    \multicolumn{1}{r}{} &       & \multicolumn{4}{c|}{PredCls} & \multicolumn{4}{c|}{SGCls} & \multicolumn{4}{c}{SGDet} \\
    \multicolumn{1}{r}{} &       & mR@20    & mR@50    & mR@100   & \multicolumn{1}{l|}{$\text {AVG}_{mR}$} & mR@20    & mR@50    & mR@100   & \multicolumn{1}{l|}{$\text {AVG}_{mR}$} & mR@20    & mR@50    & mR@100   & \multicolumn{1}{l}{$\text {AVG}_{mR}$} \\
    \midrule
    \midrule
    \multicolumn{2}{l|}{MotifsNet (backbone) \cite{Neuralmotifs} } & $12.2$  & $15.5$  & $16.8$  & $14.8$  & $7.2$   & $9.0$     & $9.5$   & $8.6$   & $5.2$   & $7.2$   & $8.5$   & $7.0$  \\
    \multicolumn{2}{l|}{\quad $\text {TDE \cite{TDE}}^{\Diamond \dagger}$   \tiny \textit {(CVPR’20)}} & $18.5$  & $25.5$  & $29.1$  & $24.4$  & $9.8$   & $13.1$  & $14.9$  & $12.6$  & $5.8$   & $8.2$   & $9.8$   & $7.9$  \\
    \multicolumn{2}{l|}{\quad $\text {SegG \cite{SegG}}^{\Delta \dagger}$  \tiny \textit {(ICCV’21)} } & $14.5$  & $18.5$  & $20.2$  & $17.7$  & $8.9$   & $11.2$  & $12.1$  & $10.7$  & $6.4$   & $8.3$   & $9.2$   & $8.0$  \\
    \multicolumn{2}{l|}{\quad $\text {BPL+SA \cite{BPLSA}}^{\blacklozenge \Diamond \dagger}$  \tiny \textit {(ICCV’21)} } & $24.8$  & $29.7$  & $31.7$  & $28.7$  & $14.0$    & $16.5$  & $17.5$  & $16.0$  & $10.7$  & $13.5$  & $15.6$  & $13.3$  \\
    \multicolumn{2}{l|}{\quad $\text {CogTree \cite{Cogtree}}^{\blacklozenge \dagger}$  \tiny \textit {(IJCAI’21)} } & $20.9$  & $26.4$  & $29.0$  & $25.4$  & $12.1$  & $14.9$  & $16.1$  & $14.4$  & $7.9$   & $10.4$  & $11.8$  & $10.0$  \\
    \multicolumn{2}{l|}{\quad $\text {DLFE \cite{DLFE}}^{\Diamond \dagger}$  \tiny \textit {(MM’21)} } & $22.1$  & $26.9$  & $28.8$  & $25.9$  & $12.8$  & $15.2$  & $15.9$  & $14.6$  & $8.6$   & $11.7$  & $13.8$  & $11.4$  \\
    \multicolumn{2}{l|}{\quad $\text {EBM-loss \cite{EBMloss}}^{\blacklozenge \dagger}$  \tiny \textit {(CVPR’21)} } & $14.2$  & $18.0$    & $19.5$  & $17.2$  & $8.2$   & $10.2$  & $11.0$    & $9.8$   & $5.7$   & $7.7$   & $9.3$   & $7.6$  \\
    \multicolumn{2}{l|}{\quad $\text {Loss-reweight \cite{logitadjustment}}^{\blacklozenge \dagger}$   \tiny \textit {(ICLR’21)} } & $26.5$  & $32.9$  & $35.3$  & $31.6$  & $13.8$  & $17.4$  & $19.3$  & $16.8$  & $9.2$  & $12.8$  & $16.5$  & $12.8$   \\
    \multicolumn{2}{l|}{\quad $\text {Logit-reweight \cite{logitadjustment}}^{\Diamond \dagger}$   \tiny \textit {(ICLR’21)} } & $12.2$  & $15.4$  & $16.7$  & $14.8$  & $6.4$  & $7.6$  & $8.3$  & $7.4$  & $4.5$  & $5.9$  & $7.7$  & $6.0$   \\
    \multicolumn{2}{l|}{\quad $\text {HML \cite{HML}}^{\Delta \blacklozenge \dagger}$  \tiny \textit {(ECCV’22)} } & $30.1$  & $36.3$  & $38.7$  & $35.0$  & $17.1$  & $20.8$  & $22.1$  & $20.0$  & $10.8$  & $14.6$  & $17.3$  & $14.2$  \\
    \multicolumn{2}{l|}{\quad $\text {FGPL \cite{FGPL}}^{\blacklozenge \dagger}$  \tiny \textit {(CVPR’22)} } & $24.3$  & $33.0$  & $37.5$  & $31.6$  & $17.1$  & $21.3$  & $22.5$  & $20.3$  & $11.1$  & $15.4$  & $18.2$  & $14.9$  \\
    \multicolumn{2}{l|}{\quad $\text {TransRwt \cite{TransRwt}}^{\Delta \dagger}$  \tiny \textit {(ECCV’22)} } & $-$     & $35.8$  & $39.1$  & $-$  & $-$     & $21.$5  & $22.8$  & $-$  & $-$     & $15.8$  & $18.0$  & $-$  \\
    \multicolumn{2}{l|}{\quad $\text {GCL \cite{GCL}}^{\blacklozenge \ddagger}$  \tiny \textit {(CVPR’22)} } & $30.5$  & $36.1$  & $38.2$  & $34.9$  & $18.0$    & $20.8$  & $21.8$  & $20.2$  & $12.9$  & $16.8$  & $19.3$  & $16.3$  \\
    \multicolumn{2}{l|}{\quad $\text {PPDL \cite{PPDL}}^{\blacklozenge \dagger}$  \tiny \textit {(CVPR’22)} } & $-$     & $32.2$  & $33.3$  & $-$  & $-$     & $17.5$  & $18.2$  & $-$ & $-$     & $11.4$  & $13.5$  & $-$  \\
    \multicolumn{2}{l|}{\quad $\text {RTPB \cite{RTPB}}^{\blacklozenge \Diamond \ddagger}$  \tiny \textit {(AAAI’22)} } & $28.8$  & $35.3$  & $37.7$  & $33.9$  & $16.3$  & $19.4$  & $22.6$  & $19.4$  & $9.7$   & $13.1$  & $15.5$  & $12.8$  \\
    \multicolumn{2}{l|}{\quad $\text {NICE \cite{NICE}}^{\Delta \blacklozenge \dagger}$  \tiny \textit {(CVPR’22)} } & $-$     & $30.0$    & $32.1$  & $-$  & $-$     & $16.4$  & $17.5$  & $-$  & $-$     & $10.4$  & $12.7$  & $-$  \\
    \multicolumn{2}{l|}{\quad $\text {PKO \cite{PKO}}^{\Diamond \dagger}$  \tiny \textit {(arXiv’22)} } & $25.0$    & $31.4$  & $34.0$    & $30.1$  & $14.1$  & $17.6$  & $19.1$  & $16.9$  & $9.6$  & $13.4$  & $16.1$  & $13.0$  \\
    \multicolumn{2}{l|}{\quad $\text {LS-KD(Iter) \cite{LSKD}}^{\blacklozenge \dagger}$  \tiny \textit {(arXiv’22)} } & $-$     & $24.1$  & $27.4$  & $-$  & $-$     & $13.8$  & $15.2$  & $-$  & $-$     & $9.7$   & $11.5$  & $-$  \\
    \multicolumn{2}{l|}{\quad $\text {CAME \cite{CAME}}^{\Delta \blacklozenge \ddagger}$  \tiny \textit {(arXiv’22)} } & $18.1$  & $26.2$  & $32.0$    & $25.4$  & $10.5$  & $15.1$  & $18.0$    & $14.5$  & $6.7$   & $9.3$  & $12.1$  & $9.4$  \\
    \midrule
    \midrule
    \multicolumn{2}{l|}{\quad $\text {TsCM}^{\blacklozenge \Diamond \dagger}$} & $31.8$  & $37.8$  & $40.9$  & $36.8$  & $18.7$  & $22.4$  & $23.8$  & $21.6$  & $13.7$  & $17.4$  & $19.7$  & $16.9$   \\
    \toprule
    \end{tabular}}%
    \end{threeparttable}
    \begin{tablenotes}
    \footnotesize
    \item[\textit{Note}:] { $\text{AVG}_\text{mR}$ is the average of mR@20, mR@50, mR@100. $\Delta$,  $\blacklozenge$, and $\Diamond$ indicate resampling methods, reweighting methods, and adjustment methods, respectively. $\dagger$ and $\ddagger$ indicate model-agnostic and model-dependent, respectively.}
    \end{tablenotes}
  \label{tab:MotifsNet-mR}
  \vspace{-0.3cm}
\end{table*}%

\section{Experiments}
\subsection{Implementation}
\textit {Datasets.}
We evaluate our method on VG150 \cite{VG150}, a subset of the VG dataset \cite{VG} that includes the most frequent 150 object categories and 50 relationship classes. VG150 has about 94k images, and we follow the split in \cite{TDE}, \ie, 62k training images, 5k validation images, and 26k test images.

\textit {Evaluation modes.}
Following MotifsNet \cite{Neuralmotifs}, we use three evaluation modes: 1) Predicate classification (PredCls). This mode requires the SGG model to predict relationships given the ground truth boxes and object classes. 2) Scene Graph Classification (SGCls). This mode requires the SGG model to predict object classes and relationships given the ground truth boxes. 3) Scene Graph Detection (SGDet). This mode requires the SGG model to predict object classes, boxes, and relationships.

\textit {Evaluation metrics.} Following \cite{VG150,VCtree,KERN}, we adopt three evaluation metrics: 1) Recall rate (R@K). R@K is one of the most commonly used evaluation metrics, which calculates the fraction of times the correct relationship is predicted in the top K confident relationship predictions. Typically, K is set to 20, 50, and 100, \ie, R@20, R@50, and R@100. 2) mean recall rate (mR@K). mR@K calculates the mean of the R@K for each relationship. Compared with R@K, mR@K can comprehensively evaluate the model performance on all relationship categories, especially the tail relationships. 3) Mean of R@K and mR@K (MR@K). Due to the severely long-tailed distribution, the SGG model only needs to perform well on a few head categories to achieve high R@K. Although some current works can achieve a high mR@K, they greatly sacrifice the R@K of the head categories, which is certainly not what we expected since the head categories account for significant proportions in realistic scenarios. We therefore aim to achieve a favorable tradeoff between R@K and mR@K, allowing the model to accommodate both head and tail relationships, which in turn enhances the practical value of the generated scene graph. For this purpose, we calculate the mean of R@K and mR@K, denoted as MR@K, to evaluate the model comprehensively. 

\begin{table*}[htbp]
  \centering
  \caption{Quantitative results (mR@K) of our method and other baselines on the VCTree backbone}
  \begin{threeparttable}
  \vspace{-0.4cm}
    \setlength\tabcolsep{2pt}
\setlength\arrayrulewidth{0.2mm}
    \resizebox{17cm}{!}{\begin{tabular}{l|l|cccc|cccc|cccc}
    \toprule
    \multicolumn{1}{r}{} &       & \multicolumn{4}{c|}{PredCls} & \multicolumn{4}{c|}{SGCls} & \multicolumn{4}{c}{SGDet} \\
    \multicolumn{1}{r}{} &       & mR@20    & mR@50    & mR@100   & \multicolumn{1}{l|}{$\text {AVG}_{mR}$} & mR@20    & mR@50    & mR@100   & \multicolumn{1}{l|}{$\text {AVG}_{mR}$} & mR@20    & mR@50    & mR@100   & \multicolumn{1}{l}{$\text {AVG}_{mR}$} \\
    \midrule
    \midrule
    \multicolumn{2}{l|}{VCTree (backbone) \cite{VCtree}} &$12.4$ 	&$15.4$ 	&$16.6$ 	&$14.8$ 	&$6.3$ 	&$7.5$ 	&$8.0$ 	&$7.3$ 	&$4.9$ 	&$6.6$ 	&$7.7$ 	&$6.4$   \\
    \multicolumn{2}{l|}{\quad $\text {TDE \cite{TDE}}^{\Diamond \dagger}$   \tiny \textit {(CVPR’20)} } &$18.4$ 	&$25.4$ 	&$28.7$ 	&$24.2$ 	&$8.9$ 	&$12.2$ 	&$14.0$ 	&$11.7$ 	&$6.9$ 	&$9.3$ 	&$11.1$ 	&$9.1$   \\
    \multicolumn{2}{l|}{\quad $\text {SegG \cite{SegG}}^{\Delta \dagger}$  \tiny \textit {(ICCV’21)} } &$15.0$ 	&$19.2$ 	&$21.1$ 	&$18.4$ 	&$9.3$ 	&$11.6$ 	&$12.3$ 	&$11.1$ 	&$6.3$ 	&$8.1$ 	&$9.0$ 	&$7.8$  \\
    \multicolumn{2}{l|}{\quad $\text {BPL+SA \cite{BPLSA}}^{\blacklozenge \Diamond \dagger}$  \tiny \textit {(ICCV’21)} } &$26.2 	$&$30.6 	$&$32.6 	$&$29.8 	$&$17.2 	$&$20.1 	$&$21.2 	$&$19.5 	$&$10.6 	$&$13.5 	$&$15.7 	$&$13.3$   \\
    \multicolumn{2}{l|}{\quad $\text {CogTree \cite{Cogtree}}^{\blacklozenge \dagger}$  \tiny \textit {(IJCAI’21)} } &$22.0 	$&$27.6 	$&$29.7 	$&$26.4 	$&$15.4 	$&$18.8 	$&$19.9 	$&$18.0 	$&$7.8 	$&$10.4 	$&$12.1 	$&$10.1$  \\
    \multicolumn{2}{l|}{\quad $\text {DLFE \cite{DLFE}}^{\Diamond \dagger}$  \tiny \textit {(MM’21)} } &$20.8 	$&$25.3 	$&$27.1 	$&$24.4 	$&$15.8 	$&$18.9 	$&$20.0 	$&$18.2 	$&$8.6 	$&$11.8 	$&$13.8 	$&$11.4$   \\
    \multicolumn{2}{l|}{\quad $\text {EBM-loss \cite{EBMloss}}^{\blacklozenge \dagger}$  \tiny \textit {(CVPR’21)}} &$14.2 	$&$18.2 	$&$19.7 	$&$17.4 	$&$10.4 	$&$12.5 	$&$13.5 	$&$12.1 	$&$5.7 	$&$7.7 	$&$9.1 	$&$7.5$   \\
    \multicolumn{2}{l|}{\quad $\text {Loss-reweight \cite{logitadjustment}}^{\blacklozenge \dagger}$   \tiny \textit {(ICLR’21)} } &$ 25.5  $&$ 31.7  $&$ 34.9  $&$ 30.7  $&$ 16.3  $&$ 20.9  $&$ 22.3  $&$ 19.8  $&$ 8.7  $&$ 12.4  $&$ 14.2  $&$ 11.8$   \\
    \multicolumn{2}{l|}{\quad $\text {Logit-reweight \cite{logitadjustment}}^{\Diamond \dagger}$   \tiny \textit {(ICLR’21)} } &$ 11.7  $&$ 15.2  $&$ 16.3  $&$ 14.4  $&$ 6.4  $&$ 7.6  $&$ 8.1  $&$ 7.4  $&$ 4.1  $&$ 5.5  $&$ 7.1  $&$ 5.6$   \\
    \multicolumn{2}{l|}{\quad $\text {HML \cite{HML}}^{\Delta \blacklozenge \dagger}$  \tiny \textit {(ECCV’22)} } &$31.0 	$&$36.9 	$&$39.2 	$&$35.7 	$&$20.5 	$&$25.0 	$&$26.8 	$&$24.1 	$&$10.1 	$&$13.7 	$&$16.3 	$&$13.4$  \\
    \multicolumn{2}{l|}{\quad $\text {FGPL \cite{FGPL}}^{\blacklozenge \dagger}$  \tiny \textit {(CVPR’22)} } &$30.8 	$&$37.5 	$&$40.2 	$&$36.2 	$&$21.9 	$&$26.2 	$&$27.6 	$&$25.2 	$&$11.9 	$&$16.2 	$&$19.1 	$&$15.7$   \\
    \multicolumn{2}{l|}{\quad $\text {TransRwt \cite{TransRwt}}^{\Delta \dagger}$  \tiny \textit {(ECCV’22)}} &$- 	$&$37.0 	$&$39.7 	$&$- 	$&$- 	$&$19.9 	$&$21.8 	$&$- 	$&$- 	$&$12.0 	$&$14.9 	$&$-$   \\
    \multicolumn{2}{l|}{\quad $\text {GCL \cite{GCL}}^{\blacklozenge \ddagger}$  \tiny \textit {(CVPR’22)} } &$31.4 	$&$37.1 	$&$39.1 	$&$35.9 	$&$19.5 	$&$22.5 	$&$23.5 	$&$21.8 	$&$11.9 	$&$15.2 	$&$17.5 	$&$14.9$   \\
    \multicolumn{2}{l|}{\quad $\text {PPDL \cite{PPDL}}^{\blacklozenge \dagger}$  \tiny \textit {(CVPR’22)}} &$- 	$&$33.3 	$&$33.8 	$&$- 	$&$- 	$&$21.8 	$&$22.4 	$&$- 	$&$- 	$&$11.3 	$&$13.3 	$&$-$  \\
    \multicolumn{2}{l|}{\quad $\text {RTPB \cite{RTPB}}^{\blacklozenge \Diamond \ddagger}$  \tiny \textit {(AAAI’22)} } &$27.3 	$&$33.4 	$&$35.6 	$&$32.1 	$&$20.6 	$&$24.5 	$&$25.8 	$&$23.6 	$&$9.6 	$&$12.8 	$&$15.1 	$&$12.5$  \\
    \multicolumn{2}{l|}{\quad $\text {NICE \cite{NICE}}^{\Delta \blacklozenge \dagger}$  \tiny \textit {(CVPR’22)}} &$- 	$&$30.9 	$&$33.1 	$&$- 	$&$- 	$&$20.0 	$&$21.2 	$&$- 	$&$- 	$&$10.1 	$&$12.1 	$&$-$   \\
    \multicolumn{2}{l|}{\quad $\text {PKO \cite{PKO}}^{\Diamond \dagger}$  \tiny \textit {(arXiv’22)} } &$26.1 	$&$32.2 	$&$34.6 	$&$31.0 	$&$18.4 	$&$22.3 	$&$23.7 	$&$21.5 	$&$9.6 	$&$13.2 	$&$15.9 	$&$12.9$   \\
    \multicolumn{2}{l|}{\quad $\text {LS-KD(Iter) \cite{LSKD}}^{\blacklozenge \dagger}$  \tiny \textit {(arXiv’22)} } &$- 	$&$24.2 	$&$27.1 	$&$- 	$&$- 	$&$17.3 	$&$19.1 	$&$- 	$&$- 	$&$9.7 	$&$11.3 	$&$-$  \\
    \multicolumn{2}{l|}{\quad $\text {CAME \cite{CAME}}^{\Delta \blacklozenge \ddagger}$  \tiny \textit {(arXiv’22)} } &$18.9 	$&$26.6 	$&$32.0 	$&$25.8 	$&$11.7 	$&$17.0 	$&$20.5 	$&$16.4 	$&$5.9 	$&$8.7 	$&$10.8 	$&$8.5$  \\
    \midrule
    \midrule
    \multicolumn{2}{l|}{\quad $\text {TsCM}^{\blacklozenge \Diamond \dagger}$} &$ 32.3  $&$ 38.7  $&$ 41.5  $&$ 37.5  $&$ 23.4  $&$ 26.9  $&$ 28.9  $&$ 26.4  $&$ 12.5  $&$ 16.9  $&$ 19.3  $&$ 16.2$  \\
    \toprule
    \end{tabular}}%
    \end{threeparttable}
    \begin{tablenotes}
    \footnotesize
    \item[\textit{Note}:] { $\text {AVG}_\text{mR}$ is the average of mR@20, mR@50, mR@100. $\Delta$,  $\blacklozenge$, $\Diamond$, $\dagger$ and $\ddagger$ are with the same meanings as in Table~\ref{tab:MotifsNet-mR}.}
    \end{tablenotes}
  \label{tab:VCTree-mR}%
  \vspace{-0.3cm}
\end{table*}%

\begin{table*}[htbp]
  \centering
  \caption{Quantitative results (mR@K) of our method and other baselines on the Transformer backbone}
  \begin{threeparttable}
  \vspace{-0.4cm}
    \setlength\tabcolsep{2pt}
\setlength\arrayrulewidth{0.2mm}
    \resizebox{17cm}{!}{\begin{tabular}{l|l|cccc|cccc|cccc}
    \toprule
    \multicolumn{1}{r}{} &       & \multicolumn{4}{c|}{PredCls} & \multicolumn{4}{c|}{SGCls} & \multicolumn{4}{c}{SGDet} \\
    \multicolumn{1}{r}{} &       & mR@20    & mR@50    & mR@100   & \multicolumn{1}{l|}{$\text {AVG}_{mR}$} & mR@20    & mR@50    & mR@100   & \multicolumn{1}{l|}{$\text {AVG}_{mR}$} & mR@20    & mR@50    & mR@100   & \multicolumn{1}{l}{$\text {AVG}_{mR}$} \\
    \midrule
    \midrule
    \multicolumn{2}{l|}{Transformer (backbone) \cite{transformer}} &$ 12.4  $&$ 16.0  $&$ 17.5  $&$ 15.3  $&$ 7.7   $&$ 9.6     $&$ 10.2   $&$ 9.2   $&$ 5.3   $&$ 7.3   $&$ 8.8   $&$ 7.1$  \\
    \multicolumn{2}{l|}{\quad $\text {BPL+SA \cite{BPLSA}}^{\blacklozenge \Diamond \dagger}$  \tiny \textit {(ICCV’21)} } &$ 24.8  $&$ 29.7  $&$ 31.7  $&$ 28.7  $&$ 14.0    $&$ 16.5  $&$ 17.5  $&$ 16.0  $&$ 10.7  $&$ 13.5  $&$ 15.6  $&$ 13.3$  \\
    \multicolumn{2}{l|}{\quad $\text {CogTree \cite{Cogtree}}^{\blacklozenge \dagger}$  \tiny \textit {(IJCAI’21)}} &$ 20.9  $&$ 26.4  $&$ 29.0  $&$ 25.4  $&$ 12.1  $&$ 14.9  $&$ 16.1  $&$ 14.4  $&$ 7.9   $&$ 10.4  $&$ 11.8  $&$ 10.0$  \\
    \multicolumn{2}{l|}{\quad $\text {Loss-reweight \cite{logitadjustment}}^{\blacklozenge \dagger}$   \tiny \textit {(ICLR’21)} } &$ 27.8  $&$ 33.1  $&$ 36.2  $&$ 32.4  $&$ 15.8  $&$ 19.3  $&$ 21.1  $&$ 18.8  $&$ 11.7  $&$ 15.3  $&$ 17.9  $&$ 15.0$  \\
    \multicolumn{2}{l|}{\quad $\text {Logit-reweight \cite{logitadjustment}}^{\Diamond \dagger}$   \tiny \textit {(ICLR’21)} } &$ 13.5  $&$ 16.6  $&$ 18.9  $&$ 16.3  $&$ 6.7  $&$ 8.6  $&$ 9.8  $&$ 8.4  $&$ 6.6  $&$ 8.3  $&$ 9.4  $&$ 8.1$   \\
    \multicolumn{2}{l|}{\quad $\text {HML \cite{HML}}^{\Delta \blacklozenge \dagger}$  \tiny \textit {(ECCV’22)}} &$ 30.1  $&$ 36.3  $&$ 38.7  $&$ 35.0  $&$ 17.1  $&$ 20.8  $&$ 22.1  $&$ 20.0  $&$ 10.8  $&$ 14.6  $&$ 17.3  $&$ 14.2$  \\
    \multicolumn{2}{l|}{\quad $\text {FGPL \cite{FGPL}}^{\blacklozenge \dagger}$  \tiny \textit {(CVPR’22)} } &$ 24.3  $&$ 33.0  $&$ 37.5  $&$ 31.6  $&$ 17.1  $&$ 21.3  $&$ 22.5  $&$ 20.3  $&$ 11.1  $&$ 15.4  $&$ 18.2  $&$ 14.9$  \\
    \multicolumn{2}{l|}{\quad $\text {TransRwt \cite{TransRwt}}^{\Delta \dagger}$  \tiny \textit {(ECCV’22)} } &$ -     $&$ 35.8  $&$ 39.1  $&$ -  $&$ -     $&$ 21.5  $&$ 22.8  $&$ -  $&$ -     $&$ 15.8  $&$ 18.0  $&$ -$  \\
    \midrule
    \midrule
    \multicolumn{2}{l|}{\quad $\text {TsCM}^{\blacklozenge \Diamond \dagger}$} &$ 32.8  $&$ 40.1  $&$ 42.3  $&$ 38.4  $&$ 19.6  $&$ 23.7  $&$ 25.1  $&$ 22.8  $&$ 13.8  $&$ 18.3  $&$ 21.2  $&$ 17.8$  \\
    \toprule
    \end{tabular}}%
    \end{threeparttable}
    \begin{tablenotes}
    \footnotesize
    \item[\textit{Note}:] { $\text {AVG}_\text{mR}$ is the average of mR@20, mR@50, mR@100. $\Delta$,  $\blacklozenge$, $\Diamond$, $\dagger$ and $\ddagger$ are with the same meanings as in Table~\ref{tab:MotifsNet-mR}.}
    \end{tablenotes}
  \label{tab:Transformer}%
  \vspace{-0.3cm}
\end{table*}%

\begin{table*}[htbp]
  \centering
  \caption{Quantitative results (R@K) of our method and other baselines on the MotifsNet, VCTree, and Transformer backbones}
  \begin{threeparttable}
  \vspace{-0.4cm}
      \setlength\tabcolsep{2pt}
     \setlength\arrayrulewidth{0.2mm}
    \resizebox{14cm}{!}{\begin{tabular}{l|l|cccc|cccc|cccc}
    \toprule
    \multicolumn{1}{r}{} &       & \multicolumn{4}{c|}{PredCls} & \multicolumn{4}{c|}{SGCls} & \multicolumn{4}{c}{SGDet} \\
    \multicolumn{1}{r}{} &       & R@20    & R@50    & R@100   & \multicolumn{1}{l|}{$\text {AVG}_{R}$} & R@20    & R@50    & R@100   & \multicolumn{1}{l|}{$\text {AVG}_{R}$} & R@20    & R@50    & R@100   & \multicolumn{1}{l}{$\text {AVG}_{R}$} \\
    \midrule
    \midrule
    \multicolumn{2}{l|}{MotifsNet (backbone) \cite{Neuralmotifs}} &$59.5	$&$66.0	$&$67.9	$&$64.5 	$&$35.8	$&$39.1	$&$39.9	$&$38.3 	$&$25.1	$&$32.1	$&$36.9	$&$31.4$   \\
    \multicolumn{2}{l|}{\quad $\text {TDE \cite{TDE}}^{\Diamond \dagger}$   \tiny \textit {(CVPR’20)}} &$33.6	$&$46.2	$&$51.4	$&$43.7 	$&$21.7	$&$27.7	$&$29.9	$&$26.4 	$&$12.4	$&$16.9	$&$20.3	$&$16.5$    \\
    \multicolumn{2}{l|}{\quad $\text {CogTree \cite{Cogtree}}^{\blacklozenge \dagger}$  \tiny \textit {(IJCAI’21)}} &$31.1	$&$35.6	$&$36.8 	$&$34.5 	$&$19.4	$&$21.6	$&$22.2	$&$21.1 	$&$15.7	$&$20.0	$&$22.1	$&$19.3$   \\
    \multicolumn{2}{l|}{\quad $\text {Loss-reweight \cite{logitadjustment}}^{\blacklozenge \dagger}$   \tiny \textit {(ICLR’21)} } &$ 31.4  $&$ 38.3  $&$ 40.4  $&$ 36.7  $&$ 19.9  $&$ 22.3  $&$ 24.7  $&$ 22.3  $&$ 14.8  $&$ 19.3  $&$ 21.4  $&$ 18.5$  \\
    \multicolumn{2}{l|}{\quad $\text {Logit-reweight \cite{logitadjustment}}^{\Diamond \dagger}$   \tiny \textit {(ICLR’21)} } &$ 54.8  $&$ 61.3  $&$ 63.1  $&$ 59.8  $&$ 32.3  $&$ 35.4  $&$ 36.6  $&$ 34.8  $&$ 22.3  $&$ 28.9  $&$ 34.1  $&$ 28.4$   \\
    \multicolumn{2}{l|}{\quad $\text {TransRwt \cite{TransRwt}}^{\Delta \dagger}$  \tiny \textit {(ECCV’22)}} &$-	$&$48.6	$&$50.5	$&$- 	$&$-	$&$29.4	$&$30.2	$&$- 	$&$-	$&$23.5	$&$27.2 	$&$-$  \\
    \multicolumn{2}{l|}{\quad $\text {PPDL \cite{PPDL}}^{\blacklozenge \dagger}$  \tiny \textit {(CVPR’22)}  } &$- 	$&$47.2 	$&$47.6 	$&$- 	$&$- 	$&$28.4 	$&$29.3 	$&$- 	$&$- 	$&$21.2 	$&$23.9 	$&$-$  \\
    \cdashline{1-14}[3pt/5pt]
    \multicolumn{2}{l|}{\quad $\text {TsCM}^{\blacklozenge \Diamond \dagger}$} &$ 49.7  $&$ 57.1  $&$ 59.5  $&$ 55.4  $&$ 29.8  $&$ 33.6  $&$ 34.2  $&$ 32.5  $&$ 23.7  $&$ 28.6  $&$ 29.6  $&$ 27.3$ \\
    \midrule
    \midrule
    \multicolumn{2}{l|}{VCTree (backbone) \cite{VCtree}} &$59.8	$&$66.2	$&$68.1	$&$64.7 	$&$37.0	$&$40.5	$&$41.4	$&$39.6 	$&$24.7	$&$31.5	$&$36.2	$&$30.8$    \\
    \multicolumn{2}{l|}{\quad $\text {TDE \cite{TDE}}^{\Diamond \dagger}$   \tiny \textit {(CVPR’20)}} &$39.1	$&$49.9	$&$54.5	$&$47.8 	$&$22.8	$&$28.8	$&$31.2	$&$27.6 	$&$14.3	$&$19.6	$&$23.3	$&$19.1$    \\
    \multicolumn{2}{l|}{\quad $\text {CogTree \cite{Cogtree}}^{\blacklozenge \dagger}$  \tiny \textit {(IJCAI’21)}} &$39.0	$&$44.0	$&$45.4	$&$42.8 	$&$27.8	$&$30.9	$&$31.7	$&$30.1 	$&$14.0	$&$18.2	$&$20.4	$&$17.5$   \\
    \multicolumn{2}{l|}{\quad $\text {Loss-reweight \cite{logitadjustment}}^{\blacklozenge \dagger}$   \tiny \textit {(ICLR’21)} } &$ 29.7  $&$ 36.8  $&$ 39.3  $&$ 35.3  $&$ 18.3  $&$ 21.8  $&$ 24.2  $&$ 21.4  $&$ 13.2  $&$ 17.6  $&$ 20.9  $&$ 17.2$  \\
    \multicolumn{2}{l|}{\quad $\text {Logit-reweight \cite{logitadjustment}}^{\Diamond \dagger}$   \tiny \textit {(ICLR’21)} } &$ 59.6  $&$ 64.1  $&$ 65.3  $&$ 63.0  $&$ 36.7  $&$ 38.5  $&$ 39.2  $&$ 38.1  $&$ 22.7  $&$ 29.5  $&$ 33.8  $&$ 28.7$   \\
    \multicolumn{2}{l|}{\quad $\text {TransRwt \cite{TransRwt}}^{\Delta \dagger}$  \tiny \textit {(ECCV’22)}} &$-	$&$48.0	$&$49.9	$&$- 	$&$-	$&$30.0	$&$30.9	$&$- 	$&$-	$&$23.6	$&$27.8	$&$-$  \\
    \multicolumn{2}{l|}{\quad $\text {PPDL \cite{PPDL}}^{\blacklozenge \dagger}$  \tiny \textit {(CVPR’22)}} &$-	$&$47.6	$&$48	$&$- 	$&$-	$&$32.1	$&$33	$&$- 	$&$-	$&$20.1	$&$22.9	$&$-$  \\
    \cdashline{1-14}[3pt/5pt]
    \multicolumn{2}{l|}{\quad $\text {TsCM}^{\blacklozenge \Diamond \dagger}$} &$52.6  $&$ 57.5  $&$ 60.3  $&$ 56.8  $&$ 30.8  $&$ 33.7  $&$ 35.2  $&$ 33.2 $&$ 25.3  $&$ 29.2  $&$ 30.4  $&$ 28.3$   \\
    \midrule
    \midrule
    \multicolumn{2}{l|}{Transformer (backbone) \cite{transformer}} &$58.5	$&$65.0	$&$66.7	$&$63.4 	$&$35.6	$&$38.9	$&$39.8	$&$38.1 	$&$24.0	$&$30.3	$&$33.3	$&$29.2$  \\
    \multicolumn{2}{l|}{\quad $\text {CogTree \cite{Cogtree}}^{\blacklozenge \dagger}$  \tiny \textit {(IJCAI’21)}} &$34.1	$&$38.4	$&$39.7	$&$37.4 	$&$20.8	$&$22.9	$&$23.4	$&$22.4 	$&$15.1	$&$19.5	$&$21.7	$&$18.8$  \\
    \multicolumn{2}{l|}{\quad $\text {Loss-reweight \cite{logitadjustment}}^{\blacklozenge \dagger}$   \tiny \textit {(ICLR’21)} } &$ 29.7  $&$ 36.2  $&$ 38.6  $&$ 34.8  $&$ 18.2  $&$ 21.4  $&$ 22.5  $&$ 20.7  $&$ 13.1  $&$ 17.5  $&$ 20.9  $&$ 17.1$  \\
    \multicolumn{2}{l|}{\quad $\text {Logit-reweight \cite{logitadjustment}}^{\Diamond \dagger}$   \tiny \textit {(ICLR’21)} } &$ 53.6  $&$ 60.4  $&$ 62.5  $&$ 58.8  $&$ 34.2  $&$ 36.8  $&$ 37.6  $&$ 36.2  $&$ 21.1  $&$ 27.7  $&$ 31.2  $&$ 26.6$   \\
    \multicolumn{2}{l|}{\quad $\text {TransRwt \cite{TransRwt}}^{\Delta \dagger}$  \tiny \textit {(ECCV’22)}} &$-	$&$49.0	$&$50.8	$&$- 	$&$-	$&$29.6	$&$30.5	$&$- 	$&$-	$&$23.1	$&$27.1	$&$-$   \\
    \cdashline{1-14}[3pt/5pt]
    \multicolumn{2}{l|}{\quad $\text {TsCM}^{\blacklozenge \Diamond \dagger}$} &$ 43.2  $&$ 53.1  $&$ 57.4  $&$ 51.2  $&$ 25.3  $&$ 29.8  $&$ 32.1  $&$ 29.1  $&$ 22.1  $&$ 26.2  $&$ 27.7  $&$ 25.3$   \\
    \toprule
    \end{tabular}}%
    \end{threeparttable}
    \begin{tablenotes}
    \footnotesize
    \item[\textit{Note}:] { $\text {AVG}_\text{R}$ is the average of R@20, R@50, R@100. $\Delta$,  $\blacklozenge$, $\Diamond$, $\dagger$ and $\ddagger$ are with the same meanings as in Table~\ref{tab:MotifsNet-mR}.}
    \end{tablenotes}
  \label{tab:recall}%
  \vspace{-0.3cm}
\end{table*}%

\textit {Training and testing.} We evaluate our model-agnostic method on the popular SGG backbones, including MotifsNet \cite{Neuralmotifs}, VCTree \cite{VCtree}, and Transformer \cite{transformer}, in the repository provided by \cite{TDE}. We follow most of the settings of this repository: 1) The object detector in the pipeline is the Faster R-CNN \cite{fasterrcnn} with the backbone of ResNeXt-101-FPN \cite{resnet}. The detector was trained with the VG training set and achieved 28.14 mAP on the VG test set. 2) The detector is then frozen and outputs the bounding boxes, categories, and features of the detected objects for the relationship classifier in the pipeline. The classifier is supervised by our proposed P-Loss and optimized by SGD. For MotifsNet \cite{Neuralmotifs} and VCTree \cite{VCtree}, the batch size and initial learning rate are set to 12 and 0.01, while these parameters in Transformer \cite{transformer} are 16 and 0.001. We set $\alpha$ in Equation (\ref{eq:PLloss}) to 5 unless otherwise mentioned. 3) In the testing phase, the logits will first be augmented by Equation (\ref{eq:logit-agument}) and then adjusted by the adjustment factors learned by Equation (\ref{eq:adjust}) to obtain the final predictions. The $\beta$ in Equation (\ref{eq:adjust_matrix}) is set to 3.

\subsection{Comparison with state-of-the-art}
\subsubsection{Backbones and baselines}
\textit {Backbones.} 
We evaluate our proposed method with three popular SGG backbones, \ie, MotifsNet \cite{Neuralmotifs}, VCTree \cite{VCtree}, and Transformer \cite{transformer}. Specifically, we first replace the loss function of the above backbones with the P-Loss to supervise the model training. We then leverage AL-Adjustment to optimize the logits outputted by the trained model during inference.

\begin{table*}[htbp]
  \centering
  \caption{Quantitative results (MR@K) of our method and other baselines on the MotifsNet, VCTree, and Transformer backbones}
  \begin{threeparttable}
  \vspace{-0.4cm}
      \setlength\tabcolsep{2pt}
     \setlength\arrayrulewidth{0.2mm}
    \resizebox{14cm}{!}{\begin{tabular}{l|ccc|ccc|ccc}
    \toprule
          & \multicolumn{3}{c|}{PredCls} & \multicolumn{3}{c|}{SGCls} & \multicolumn{3}{c}{SGDet} \\
          & \multicolumn{1}{l}{${AVG}_{mR}$} & \multicolumn{1}{l}{${AVG}_{R}$} & \multicolumn{1}{l|}{MR@K} & \multicolumn{1}{l}{${AVG}_{mR}$} & \multicolumn{1}{l}{${AVG}_{R}$} & \multicolumn{1}{l|}{MR@K} & \multicolumn{1}{l}{${AVG}_{mR}$} & \multicolumn{1}{l}{${AVG}_{R}$} & \multicolumn{1}{l}{MR@K} \\
    \midrule
    {MotifsNet (backbone) \cite{Neuralmotifs}}    &$14.8	$&$64.5	$&$39.7	$&$8.6	$&$38.3	$&$23.5	$&$7.0	$&$31.4	$&$19.2$  \\
    {\quad $\text {TDE \cite{TDE}}^{\Diamond \dagger}$   \tiny \textit {(CVPR’20)}}    &$24.4	$&$43.7	$&$34.1	$&$12.6	$&$26.4	$&$19.5	$&$7.9	$&$16.5	$&$12.2$  \\
    {\quad $\text {CogTree \cite{Cogtree}}^{\blacklozenge \dagger}$  \tiny \textit {(IJCAI’21)}}   &$25.4	$&$34.5	$&$30.0	$&$14.4	$&$21.1	$&$17.8	$&$10.0	$&$19.3	$&$14.7$  \\
    {\quad $\text {Loss-reweight \cite{logitadjustment}}^{\blacklozenge \dagger}$   \tiny \textit {(ICLR’21)} }   &$31.6	$&$36.7	$&$34.2	$&$16.8	$&$22.3	$&$19.6	$&$12.8	$&$18.5	$&$15.7$  \\
    {\quad $\text {Logit-reweight \cite{logitadjustment}}^{\Diamond \dagger}$   \tiny \textit {(ICLR’21)} }    &$14.8	$&$59.8	$&$37.3	$&$7.4	$&$34.8	$&$21.1	$&$6.0	$&$28.4	$&$17.2$  \\
    \cdashline{1-10}[3pt/5pt]
    {\quad $\text {TsCM}^{\blacklozenge \Diamond \dagger}$}    &$36.8	$&$55.4	$&$46.1	$&$21.6	$&$32.5	$&$27.1	$&$16.9	$&$27.3	$&$22.1$  \\
    \midrule
    \midrule
    {VCTree (backbone) \cite{VCtree}}    &$14.8	$&$64.7	$&$39.8	$&$7.3	$&$39.6	$&$23.5	$&$6.4	$&$30.8	$&$18.6$  \\
    {\quad $\text {TDE \cite{TDE}}^{\Diamond \dagger}$   \tiny \textit {(CVPR’20)}}    &$24.2	$&$47.8	$&$36.0	$&$11.7	$&$27.6	$&$19.7	$&$9.1	$&$19.1	$&$14.1$  \\
    {\quad $\text {CogTree \cite{Cogtree}}^{\blacklozenge \dagger}$  \tiny \textit {(IJCAI’21)}}    &$26.4	$&$42.8	$&$34.6	$&$18.0	$&$30.1	$&$24.1	$&$10.1	$&$17.5	$&$13.8$  \\
    {\quad $\text {Loss-reweight \cite{logitadjustment}}^{\blacklozenge \dagger}$   \tiny \textit {(ICLR’21)} }    &$30.7	$&$35.3	$&$33.0	$&$19.8	$&$21.4	$&$20.6	$&$11.8	$&$17.2	$&$14.5$  \\
    {\quad $\text {Logit-reweight \cite{logitadjustment}}^{\Diamond \dagger}$   \tiny \textit {(ICLR’21)} }    &$14.4	$&$63.0	$&$38.7	$&$7.4	$&$38.1	$&$22.8	$&$5.6	$&$28.7	$&$17.2$  \\
    \cdashline{1-10}[3pt/5pt]
    {\quad $\text {TsCM}^{\blacklozenge \Diamond \dagger}$}    &$37.5	$&$56.8	$&$47.2	$&$26.4	$&$33.2	$&$29.8	$&$16.2	$&$28.3	$&$22.3$  \\
    \midrule
    \midrule
    {Transformer (backbone) \cite{transformer}}   &$15.3	$&$63.4	$&$39.4	$&$9.2	$&$38.1	$&$23.7	$&$7.1	$&$29.2	$&$18.2$  \\
    {\quad $\text {CogTree \cite{Cogtree}}^{\blacklozenge \dagger}$  \tiny \textit {(IJCAI’21)}}  &$25.4	$&$37.4	$&$31.4	$&$14.4	$&$22.4	$&$18.4	$&$10.0	$&$18.8	$&$14.4$  \\
    {\quad $\text {Loss-reweight \cite{logitadjustment}}^{\blacklozenge \dagger}$   \tiny \textit {(ICLR’21)} }    &$32.4	$&$34.8	$&$33.6	$&$18.8	$&$20.7	$&$19.8	$&$15.0	$&$17.1	$&$16.1$  \\
    {\quad $\text {Logit-reweight \cite{logitadjustment}}^{\Diamond \dagger}$   \tiny \textit {(ICLR’21)} }    &$16.3	$&$58.8	$&$37.6	$&$8.4	$&$36.2	$&$22.3	$&$8.1	$&$26.6	$&$17.4$ \\
    \cdashline{1-10}[3pt/5pt]
    {\quad $\text {TsCM}^{\blacklozenge \Diamond \dagger}$}    &$38.4	$&$51.2	$&$44.8	$&$22.8	$&$29.1	$&$26.0	$&$17.8	$&$25.3	$&$21.6$  \\
    \bottomrule
    \end{tabular}}%
    \end{threeparttable}
    \begin{tablenotes}
    \footnotesize
    \item[\textit{Note}:] { \quad \quad \quad \quad MR@K is the average of $\text {AVG}_\text{mR}$ and $\text {AVG}_\text{R}$. $\Delta$, $\blacklozenge$, $\Diamond$, $\dagger$ and $\ddagger$ are with the same meanings as in Table~\ref{tab:MotifsNet-mR}.}
    \end{tablenotes}
  \label{tab:recall-meanrecall}%
  \vspace{-0.3cm}
\end{table*}%

\begin{figure*}
	\footnotesize\centering
	\centerline{\includegraphics[width=0.95\linewidth]{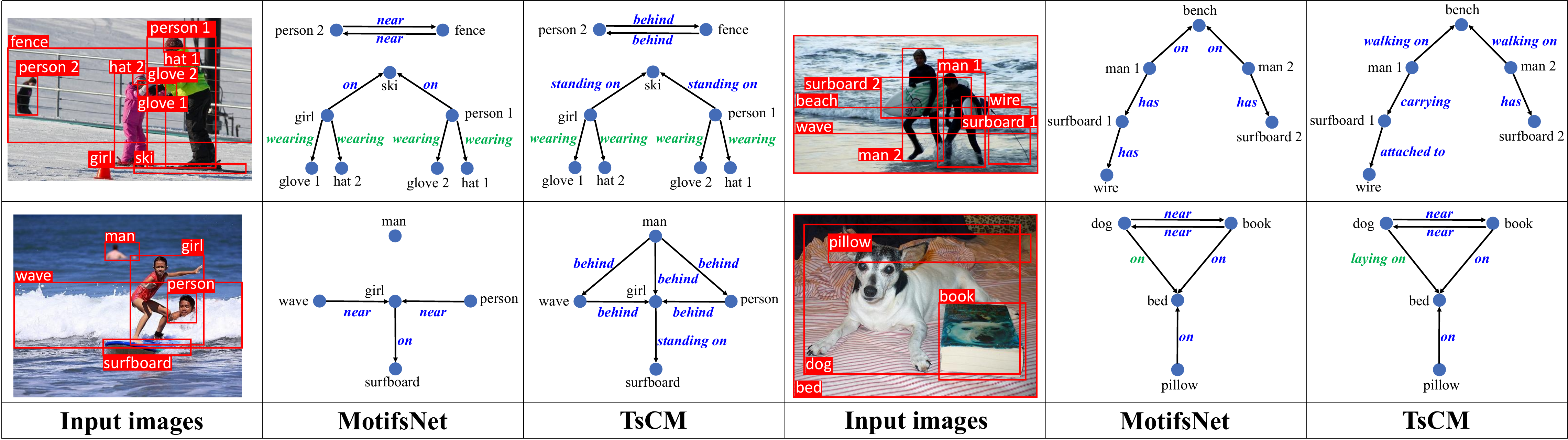}}
        \vspace{-0.2cm}
        \caption{Qualitative results of our method and baseline approach (MotifsNet \cite{Neuralmotifs}). Different relationship predictions are highlighted in blue.}
	\label{QualitativeResult}
        \vspace{-0.3cm}
\end{figure*}

\textit {Baselines.} 
We classify existing baselines from two perspectives to comprehensively evaluate our proposed framework. 1) Debiasing perspective. We divide the baselines into four groups, resampling methods, reweighting methods, adjustment methods, and hybrid methods. Resampling methods include SegG \cite{SegG} and TransRwt \cite{TransRwt}. Reweighting methods include CogTree \cite{Cogtree}, EBM-loss \cite{EBMloss}, Loss-reweight \cite{logitadjustment}, FGPL \cite{FGPL}, GCL \cite{GCL}, PPDL \cite{PPDL}, and LS-KD(Iter) \cite{LSKD}. Adjustment methods include TDE \cite{TDE}, DLFE \cite{DLFE}, Logit-reweight \cite{logitadjustment}, and PKO \cite{PKO}. Hybrid methods include BPL+SA \cite{BPLSA}, HML \cite{HML}, RTPB \cite{RTPB}, NICE \cite{NICE}, and CAME \cite{CAME}. We group from this perspective because Stage 1 in our framework is the reweighting method and Stage 2 is the adjustment method, and thus our TsCM is a hybrid method. 2) Model perspective. We divide the baselines into two groups, model-agnostic and model-dependent methods. The former group includes TDE \cite{TDE}, Loss-reweight \cite{logitadjustment}, Logit-reweight \cite{logitadjustment}, BPL+SA \cite{BPLSA}, CogTree \cite{Cogtree}, DLFE \cite{DLFE}, HML \cite{HML}, FGPL \cite{FGPL}, TransRwt \cite{TransRwt}, SegG \cite{SegG}, EBM-loss \cite{EBMloss}, PPDL \cite{PPDL}, NICE \cite{NICE}, PKO \cite{PKO}, LS-KD (Iter) \cite{LSKD}, and the latter group includes GCL \cite{GCL}, RTPB \cite{RTPB}, CAME \cite{CAME}. It is generally possible to easily transfer model-agnostic methods to different SGG backbones, thereby generalizing well in real-world applications.  

\subsubsection{Performance analysis}

\textit {Quantitative results analysis.} We report the quantitative results in Table~\ref{tab:MotifsNet-mR}, Table~\ref{tab:VCTree-mR}, Table~\ref{tab:Transformer}, Table~\ref{tab:recall}, and Table~\ref{tab:recall-meanrecall}. Our proposed method achieves state-of-the-art performance on mR@K, the most popular metric for evaluating unbiased SGG. Besides, the proposed method shows more gains on the metrics of R@K and MR@K, which indicates that TsCM obtains a better tradeoff between head and tail categories. 

From the quantitative results, we have the following observations: 1) Adjustment methods \cite{TDE,DLFE,logitadjustment,PKO} are the most relevant to our proposed AL-Adjustment approach since they share the same insight in encouraging predicting more informative tail relationships by adjusting the output logits. However, the adjustment factors in our method are adaptively learned from the observed data and thus can support causal calibration since they are sparse and independent. Benefiting from this, for instance, in PredCls mode, TsCM achieves 6.7\%/6.5\% performance gains on MotifsNet (Table~\ref{tab:MotifsNet-mR})/VCTree (Table~\ref{tab:VCTree-mR}) compared with adjustment methods. 2) Reweighting methods \cite{Cogtree,EBMloss,logitadjustment,FGPL,GCL,PPDL,LSKD} suppress partial relationships by modifying the loss function and are thus highly related to our proposed P-Loss as well. However, the difference is that our method focuses on relationships with semantic confusion, which this group of baseline methods has not explored yet. Thanks to P-Loss for providing the causal representation that can distinguish similar relationships, for example, in SGCls mode, TsCM surpasses the reweighting methods on MotifsNet (Table~\ref{tab:MotifsNet-mR})/Transformer (Table~\ref{tab:Transformer}) by 1.3\%/2.5\%. 3) Compared with hybrid methods \cite{BPLSA,HML,RTPB,NICE,CAME}, for example, in SGDet mode, our method observes 2.8\%/3.6\% improvements on VCTree (Table~\ref{tab:VCTree-mR})/Transformer (Table~\ref{tab:Transformer}). We believe this is mainly due to the fact that the two stages in our causal framework target different biases. While baseline methods mix different techniques, they only target the same bias. 4) We model the SCM with the data-level confounders so that our method is model-agnostic. TsCM can therefore be used for any SGG backbone that wants to pursue unbiased predictions. Compared with model-agnostic methods \cite{TDE,logitadjustment,logitadjustment,BPLSA,Cogtree,DLFE,HML,FGPL,TransRwt,SegG,EBMloss,PPDL,NICE,PKO,LSKD}, for instance, in SGDet mode, we observe 2.7\%/2.8\% improvements on MotifsNet (Table~\ref{tab:MotifsNet-mR})/Transformer (Table~\ref{tab:Transformer}). 5) Table~\ref{tab:recall} shows that our method is slightly weaker than logit-reweight \cite{logitadjustment} in terms of R@K. However, \cite{logitadjustment} provides biased prediction, resulting in a poor performance in mR@K, \eg, 14.8\% mR@K in the PredCls mode of the MotifsNet backbone \cite{Neuralmotifs}, while our method achieves 36.8\% in the same set-up. This indicates that our approach significantly outperforms traditional logit-adjusted methods \cite{logitadjustment} in terms of unbiased prediction. We attribute this to the inability of conventional logit-adjusted methods, particularly those employing non-heuristic prior knowledge, to effectively adjust for severely biased models in the presence of extremely long-tailed data within the SGG task. Compared with other methods, for instance, in PredCls mode, TsCM achieves 11.7\%/13.8\% gains on MotifsNet/Transformer. We believe that these exciting improvements come from sparse perturbations in our method, which do not perturb the SGG model largely, thus preserving the performance of head categories while pursuing unbiased predictions. 6) Table~\ref{tab:recall-meanrecall} shows that our method can achieve a better tradeoff between R@K and mR@K. Besides methods that benefit from recall rate (\eg, Logit-reweight \cite{logitadjustment}), our method achieves 6.4\%/7.4\%/5.4\% improvements in the backbones of MotifsNet/VCTree/Transformer. This illustrates that our method also preserves head category performance while pursuing informative tail category predictions.
\begin{figure*}
	\footnotesize\centering
	\centerline{\includegraphics[width=0.95\linewidth]{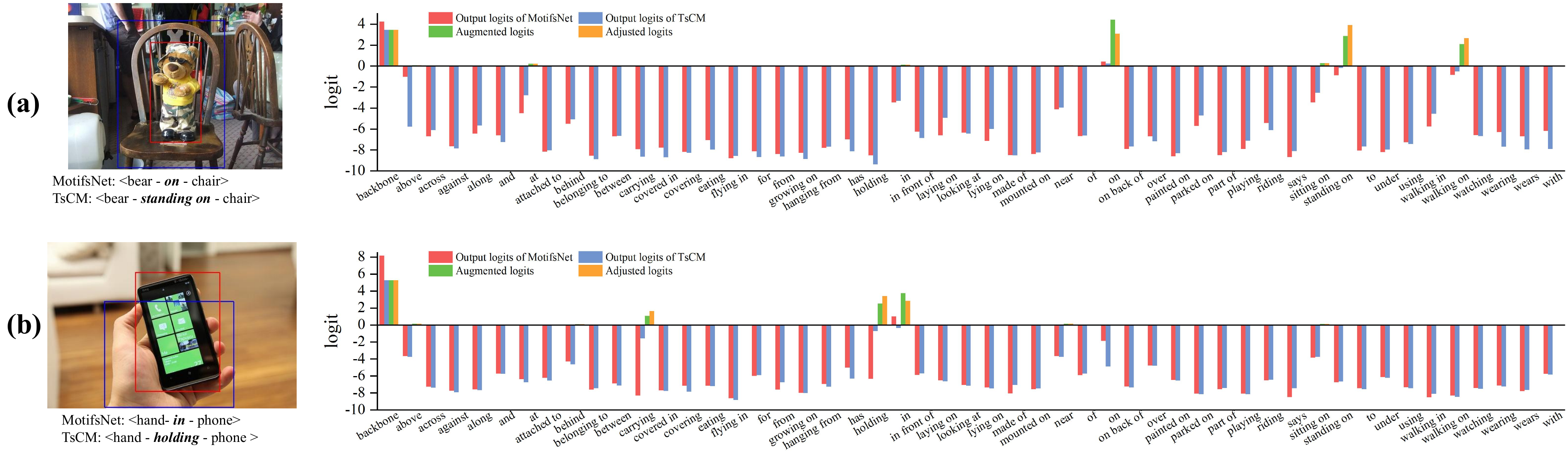}}
   \vspace{-0.2cm}
   \caption{The ablation study results of our method and baselines. Each image shows only one relationship for clarity. The red and blue bounding boxes represent subjects and objects, respectively.}
    \label{AblationResult}
    \vspace{-0.3cm}
\end{figure*}

\begin{table}[!t]
  \centering
  \caption{Results under different combinations of P-Loss and AL-Adjustment}
  \begin{threeparttable}
  \vspace{-0.4cm}
    \resizebox{90mm}{!}{\begin{tabular}{c|cc|ccc}
    \toprule
          & \multicolumn{1}{c}{} & \multicolumn{1}{c|}{} & PredCls    & SGCls    & SGDet \\
              & \multicolumn{1}{c}{P-L} & \multicolumn{1}{c|}{AL-Adj} & mR@20/50/100    & mR@20/50/100    & mR@20/50/100 \\
    \midrule
\tiny {\multirow{4}[2]{*}{\rotatebox{90}{MotifsNet}}} &   \ding{55}    &    \ding{55}   & $12.2/15.5/16.8$      &  $7.2/9.0/9.5$     & $5.2/7.2/8.5$ \\
          &    \ding{51}   &    \ding{55}   &   $12.9/16.9/20.1$    &  $7.4/9.6/11.2$     & $5.3/7.6/8.8$ \\
          &    \ding{55}   &  \ding{51}     &  $24.4/30.7/33.3$     &  $14.2/17.1/18.4$     & $8.1/10.8/13.3$ \\
          &    \ding{51}   &  \ding{51}     &   $31.8/37.8/40.9$    &  $18.7/22.4/23.8/$     & $13.7/17.4/19.7$ \\
    \midrule
   \tiny{ \multirow{4}[2]{*}{\rotatebox{90}{VCTree}}} &   \ding{55}    &    \ding{55}   &    $12.4/15.4/16.6$   &  $6.3/7.5/8.0$     & $4.9/6.6/7.7$ \\
          &    \ding{51}   &    \ding{55}   &     $12.7/16.4/19.8$  &  $8.4/10.6/11.4$     & $5.8/7.4/9.8$ \\
          &    \ding{55}   &  \ding{51}     &   $23.6/30.3/33.1$    &   $17.6/20.5/22.7$    & $10.4/13.7/15.9$ \\
          &    \ding{51}   &  \ding{51}     & $32.3/38.7/41.5$      & $23.4/26.9/28.9$      & $12.5/16.9/19.3$ \\
    \midrule
    \tiny {\multirow{4}[2]{*}{\rotatebox{90}{Transformer}}} &   \ding{55}    &  \ding{55}     &    $12.4/16.0/17.5$   &  $7.7/9.6/10.2$     & $5.3/7.3/8.8$ \\
          &    \ding{51}   &    \ding{55}   &     $13.1/17.2/20.3$  &    $9.4/11.3/12.4$   & $6.6/8.1/9.4$ \\
          &    \ding{55}   &  \ding{51}     &    $24.8/31.2/33.9$   &    $14.3/17.8/19.4$   &  $9.8/13.5/16.5$\\
          &    \ding{51}   &  \ding{51}     &   $32.8/40.1/42.3$   & $19.6/23.7/25.1$      & $13.8/18.3/21.2$ \\
    \bottomrule
    \end{tabular}}%
    \end{threeparttable}
    \begin{tablenotes}
    \footnotesize
    \item[\textit{Note}:] {P-L and AL-Adj represent the two key components of TcCM: P-Loss and \\ AL-Adjustment, respectively.}
    \end{tablenotes}
  \label{combinations}%
  \vspace{-0.3cm}
\end{table}%

\textit {Qualitative results analysis.} Fig.~\ref{QualitativeResult} shows the qualitative results generated by the original MotifsNet \cite{Neuralmotifs} and MotifsNet equipped with our TsCM. From these qualitative results, we have the following observations: 1) Our proposed method tends to predict more informative relationships, for instance, $\{$ \textit {$<$girl, standing on, ski$>$ vs $<$girl, on, ski$>$} $\}$ and $\{$ \textit {$<$dog, laying on, bed$>$ vs $<$dog, on, bed$>$} $\}$. We believe these improvements are due, in part, to the fact that our proposed AL-Adjustment can refine less informative predictions into high-informative outputs, and we will discuss this in the ablation study. 2) Our method performs well in distinguishing similar relationships, for example, $\{$ \textit {$<$wire, attached to, surfboard 1$>$ vs $<$surfboard 1, has, wire$>$} $\}$. Besides, for objects where two bounding boxes do not intersect, our method can still generate meaningful relationships, \eg, $\{$ \textit {$<$person, behind, girl$>$ vs $<$person, near, girl$>$} $\}$.  It is evident from the above improvements that our method can optimize the features of the model and classify relationships based on more than just simple information (\eg, the distance between objects). We think the optimized features are achieved by our proposed P-Loss, which can learn representations that build causality between relationships.

\begin{table}[htbp]
  \centering
  \caption{Results obtained by different logit augmentation strategies}
  \begin{threeparttable}
  \vspace{-0.4cm}
    \resizebox{85mm}{!}{\begin{tabular}{c|c|ccc}
    \toprule
          & \multicolumn{1}{c|}{Logit augmentation} & mR@20    & mR@50    & mR@100 \\
    \midrule
    
\tiny {\multirow{3}[2]{*}{\rotatebox{90}{MotifsNet}}} &   No augmentation   &$28.3$   &$34.1$   &$37.4$ \\
          &$e^{f_{\theta^*, y}(x)} \times 1$   &$30.9$   &$36.4$   &$39.7$ \\
          &$e^{f_{\theta^*, y}(x)}\times 2$   &$30.7$   &$36.5$   &$40.3$ \\
          &  $e^{f_{\theta^*, y}(x)} \times f_{\theta^*, y}^{\textrm{bg}}(x)$     &   $31.8$    &  $37.8$     & $40.9$ \\
    \midrule
\tiny {\multirow{3}[2]{*}{\rotatebox{90}{VCTree}}} &   No augmentation   &$29.6$   &$35.1$   &$38.2$ \\
          &$e^{f_{\theta^*, y}(x)}\times 1$   &31.6   &38.1   &40.8 \\
          &$e^{f_{\theta^*, y}(x)}\times 2$   &$31.4$   &$37.3$   &$40.9$ \\
          &  $e^{f_{\theta^*, y}(x)} \times f_{\theta^*, y}^{\textrm{bg}}(x)$     & $32.3$      & $38.7$      & $41.5$ \\
    \midrule
    \tiny {\multirow{3}[2]{*}{\rotatebox{90}{Transformer}}} &  No augmentation     &$30.4$   &$36.2$   &$38.7$   \\
          &$e^{f_{\theta^*, y}(x)} \times 1$   &$31.9$   &$38.4$   &$41.7$ \\
          &$e^{f_{\theta^*, y}(x)}\times 2$   &$32.2$   &$38.0$   &$41.5$ \\
          &  $e^{f_{\theta^*, y}(x)} \times f_{\theta^*, y}^{\textrm{bg}}(x)$      &   $32.8$   & $40.1$      & $42.3$ \\
    \bottomrule
    \end{tabular}}%
    \end{threeparttable}
    \begin{tablenotes}
    \footnotesize
    \item[\textit{Note}:] { The evaluation mode here is PredCls.}
    \end{tablenotes}
  \label{tab:logitaugment}%
  \vspace{-0.3cm}
\end{table}%

\subsection{Ablation Study}
\label{Ablation}
\textit {Exploring the contributions of the two stages.} TsCM consists of P-Loss and AL-Adjustment to eliminate the confounders $S$ and $L$, respectively. We first ablate our proposed causal framework using different combinations of P-Loss and AL-Adjustment, and the results are shown in Table~\ref{combinations}. The results show that both components of TsCM contribute a lot to the performance. Specifically, for AL-Adjustment, it can significantly improve the mean recall rate of the model. For example, VCTree \cite{VCtree} equipped with AL-Adjustment has 11.2\%/14.9\%/16.5\% gains on the metrics of mR@20/50/100. Although we can only observe trivial boosts for P-Loss alone, its purpose is to obtain causal representations that can well distinguish similar relationships. Therefore, these trivial boosts can be seen as by-products of the pursuit of causal representations. Table~\ref{combinations} shows that the causal representation greatly enhances AL-Adjustment. For instance, AL-Adjustment equipped with P-Loss achieves 8.7\%/8.4\%/8.4\% improvements on the backbone of VCTree \cite{VCtree}.

\begin{figure*}
	\footnotesize\centering
	\centerline{\includegraphics[width=0.95\linewidth]{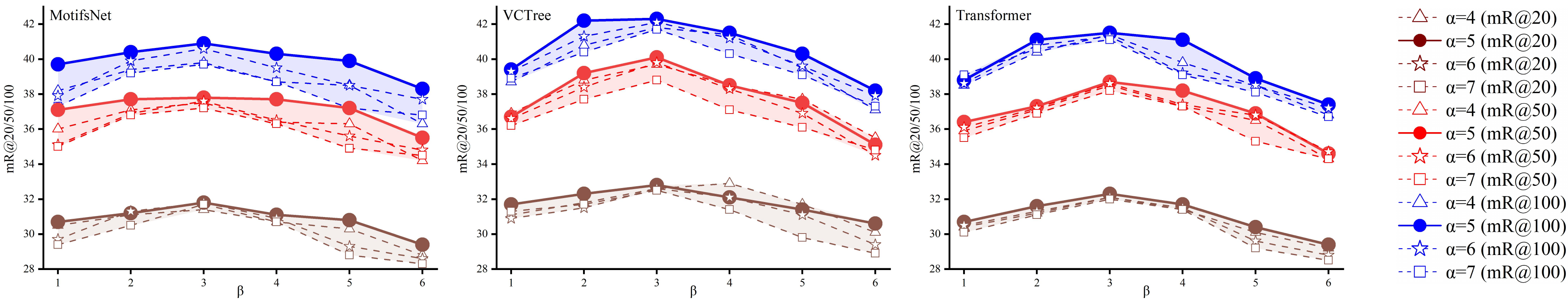}}
     \vspace{-0.2cm}
     \caption{Model performance with different $\alpha$ and $\beta$. The evaluation mode here is PredCls. The shaded areas represent the upper and lower bounds of performance for different combinations of $\alpha$ and $\beta$.}
     \label{hyperparameters}
     \vspace{-0.3cm}
\end{figure*}

We also present the output logits, the augmented logits, and the adjusted logits in Fig.~\ref{AblationResult} to show the process of P-Loss and AL-Adjustment adjusting the SGG model. These results show that AL-Adjustment can adjust less informative predictions to high-informative ones. For example, \textit {$<$bear, on, chair$>$} is adjusted to \textit {$<$bear, standing on, chair$>$} (see Fig.~\ref{AblationResult} (a)). Then, thanks to the proposed P-Loss, compared with MotifsNet \cite{Neuralmotifs}, TsCM performs better at distinguishing similar relationships, \ie, similar relationships have more significant logit gaps. As an example shown in Fig.~\ref{AblationResult} (b): In the output logits of MotifsNet \cite{Neuralmotifs}, \textit {carrying} has a 1.31× logit gap over \textit {holding}, but in TsCM, it is 2.18×. Large logit gaps will clarify the decision boundary between similar relationships, thereby overcoming semantic confusion. Finally, we can observe the issues discussed in Section \ref{sec3.3.1}, \ie, the logits of the foreground relationships being less discriminative and the logits of alternating positive and negative. It is possible, however, to unify logits to positive values and improve their discrimination, especially for the top few large logits, by using our logit augmentation method (Equation (\ref{eq:logit-agument})). We argue that the logit augmentation procedure is critical for learning adjustment factors. To prove this, we ablate our logit augmentation method and show the results in Table~\ref{tab:logitaugment}. These results demonstrate that our logit augmentation method can provide significant improvements. In addition,  the guidance term $f_{\theta^*, y}^{\textrm{bg}}(x)$ in Equation (\ref{eq:logit-agument}) also contributes a lot to the results. We think this is because the guidance term can make the augmented logits $\tilde{f}_{\theta^*, y}(x)$ more discriminative.
\begin{table}[!t]
  \centering
  \caption{The distribution (\%) of the ordering of the logits corresponding to the correct categories in the false predictions}
  \begin{threeparttable}
  \vspace{-0.4cm}
    \resizebox{90mm}{!}{\begin{tabular}{c|ccccc}
    \toprule
          & Ordering 2     & Ordering 3     & Ordering 4     & Ordering 5     & Ordering 6 \\
    \midrule
    MotifsNet & $72.5$  & $16.3$  & $7.3$   & $1.4$   & $0.8$ \\
    VCTree & $69.6$  & $14.8$  & $6.8$   & $2.7$   & $1.4$ \\
    Transformer & $68.1$  & $18.1$  & $7.9$   & $1.2$   & $0.7$ \\
    \bottomrule
    \end{tabular}}%
    \end{threeparttable}
    \begin{tablenotes}
    \footnotesize
    \item[\textit{Note}:] { Ordering $i$ means that the logit corresponding to the correct prediction is \\ ranked in the $i$ position. The evaluation mode here is PredCls.}
    \end{tablenotes}
  \label{tab:logitordering}%
  \vspace{-0.3cm}
\end{table}%

\begin{table}[!t]
  \centering
  \caption{Model performance trained with different loss functions}
  \begin{threeparttable}
  \vspace{-0.4cm}
    \resizebox{82mm}{!}{\begin{tabular}{c|c|ccc}
    \toprule
          & \multicolumn{1}{c|}{Loss} & mR/R@20    & mR/R@50    & mR/R@100 \\
    \midrule
    
\scriptsize {\multirow{5}[2]{*}{\rotatebox{90}{MotifsNet}}} &   $\ell$   &$12.2/59.5$   &$15.5/66.0$   &$16.8/67.9$ \\
          &  $\hat{\ell}^\triangleright$   &$12.3/58.8$   &$15.7/65.1$   &$17.2/66.3$ \\
          &  $\hat{\ell}^{\scaleobj{0.7}{\triangle}}$   &$12.3/58.5$   &$15.8/64.2$   &$17.5/65.9$ \\
          &  $\hat{\ell}^\triangleleft$    &$12.5/57.7$   &$16.1/63.3$   &$17.9/65.4$ \\
          &  $\hat{\ell}$     &$12.9/53.5$   &$16.9/59.4$   &$20.1/61.8$ \\
    \midrule
   \scriptsize{ \multirow{5}[2]{*}{\rotatebox{90}{VCTree}}} &    $\ell$   &$12.4/59.8$   &$15.4/66.2$   &$16.6/68.1$ \\
          &  $\hat{\ell}^\triangleright$   &$12.3/59.3$   &$15.6/65.4$   &$17.4/67.2$ \\
          &  $\hat{\ell}^{\scaleobj{0.7}{\triangle}}$   &$12.4/58.8$   &$15.6/64.3$   &$17.8/66.4$ \\
          &  $\hat{\ell}^\triangleleft$    &$12.5/58.2$   &$15.9/63.8$   &$18.2/65.9$ \\
          &  $\hat{\ell}$      &$12.7/54.3$   &$16.4/59.9$   &$19.8/62.3$ \\
    \midrule
    \scriptsize {\multirow{5}[2]{*}{\rotatebox{90}{Transformer}}} &  $\ell$     &$12.4/58.5$   &$16.0/65.0$   &$17.5/66.7$   \\
          &  $\hat{\ell}^\triangleright$    &$12.5/58.2$   &$16.2/64.2$   &$17.6/65.1$ \\
          &  $\hat{\ell}^{\scaleobj{0.7}{\triangle}}$   &$12.7/57.9$   &$16.3/63.8$   &$17.6/64.9$ \\
          &  $\hat{\ell}^\triangleleft$    &$12.8/57.4$   &$16.6/63.6$   &$18.1/63.7$ \\
          &  $\hat{\ell}$       &$13.1/50.8$   &$17.2/58.1$   &$20.3/61.2$  \\
    \bottomrule
    \end{tabular}}%
    \end{threeparttable}
    \begin{tablenotes}
    \footnotesize
    \item[\textit{Note}:] { The evaluation mode here is PredCls.}
    \end{tablenotes}
  \label{tab:differentloss}%
  \vspace{-0.3cm}
\end{table}%

\textit {Hyperparameters in TsCM.} This subsection ablates two critical hyperparameters, $\alpha$ in Equation (\ref{eq:PLloss}) and $\beta$ in Equation (\ref{eq:adjust_matrix}), that enable causal intervention on model sparsity. Fig.~\ref{hyperparameters} varies the $\alpha$ and $\beta$, and the results show that a suitable combination of hyperparameters is essential for our TsCM. We observe that the performance of the model increases when $\beta \in [1, 3]$, but decreases when $\beta > 3$. This can be explained by the facts shown in Table~\ref{tab:logitordering}: For a false prediction, the logit corresponding to the ground truth category is often ranked at the top of all logits and, in most cases, is the second largest. Therefore, most false predictions can be corrected by adjusting the first few logits. Table~\ref{tab:logitordering} shows the distribution of the ordering of the logits corresponding to the correct categories in the false predictions. Furthermore, adjusting logits that are ranked low often requires sharp adjustment factors (refer to the logits shown in Fig.~\ref{hyperparameters}), which in effect, learn a set of factors that overfit the observed data. Fig.~\ref{hyperparameters} also shows that the model performs best when $\alpha=5$. We think that when $\alpha$ is small, $\mathcal{P}_\alpha$ will miss some similar relationships, and conversely when $\alpha$ is large, many dissimilar relationships will be included in $\mathcal{P}_\alpha$. It is worth noting that $\alpha$ should be set according to observed data. In other words, for 50 relationship categories in VG150, $\alpha=5$ is the optimal choice, but $\alpha$ may have other optimal values for other datasets.

\begin{figure}
	\footnotesize\centering
	\centerline{\includegraphics[width=0.9\linewidth]{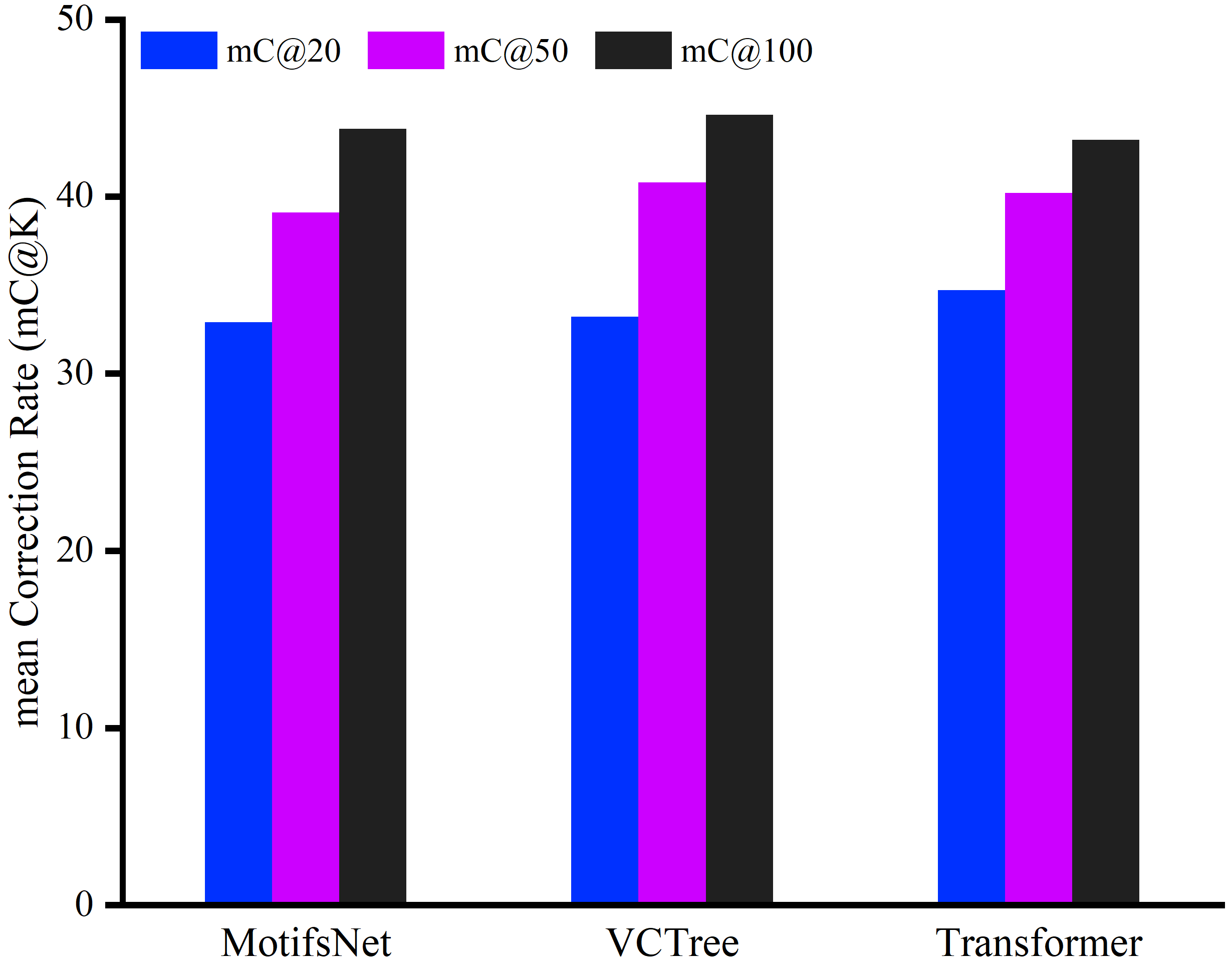}}
     \vspace{-0.2cm}
     \caption{The performance of tail categories on the proposed metric mC@K. The evaluation mode here is PredCls.}
     \label{dis_longtail}
     \vspace{-0.3cm}
\end{figure}

\textit {Disentangled claim in Equation (\ref{eq:disentangled-XYL}).} We set up three baseline loss functions to support our disentangled claims: 1) Cross-entropy loss $\ell$. 2) Modified P-Loss $\hat{\ell}^\triangleright$. This baseline loss function replaces $\mathcal{P}_{\alpha}$ in Equation (\ref{eq:PLloss}) with $\mathcal{P}_{\alpha}^\triangleright$. $\mathcal{P}_{\alpha}^\triangleright$ means to take the $\alpha$ relationships with the largest feature distance (dissimilar relationships) as relationship-population. 3) Modified P-Loss $\hat{\ell}^{\scaleobj{0.7}{\triangle}}$. This baseline loss function replaces $\mathcal{P}_{\alpha}$ in Equation (\ref{eq:PLloss}) with $\mathcal{P}_{\alpha}^{\scaleobj{0.7}{\triangle}}$. $\mathcal{P}_{\alpha}^{\scaleobj{0.7}{\triangle}}$ means to take the $\alpha^{\prime}$ relationships with the largest feature distance as relationship-population ($\alpha^{\prime}>\alpha$, $\alpha^{\prime} = 8$). 4) Modified P-Loss $\hat{\ell}^\triangleleft$. This baseline loss function replaces $\mathcal{P}_{\alpha}$ in Equation (\ref{eq:PLloss}) with $\mathcal{P}_{\alpha}^\triangleleft$. $\mathcal{P}_{\alpha}^\triangleleft$ means to take the $\alpha$ relationships with the largest feature distance belonging to the tail category as the relationship-population. Table~\ref{tab:differentloss} reports the model results supervised by different loss functions. The results of $\ell$, $\hat{\ell}^\triangleleft$, and $\hat{\ell}^\triangle$ are very close, which proves that intervening in dissimilar tail relationships has very limited impacts. Even with the possible inclusion of head categories, the supervised performance of $\hat{\ell}^\triangleright$ is still close to $\ell$ and $\hat{\ell}^\triangleleft$. This further shows that only intervening in dissimilar relationships has tiny perturbations to the model. However, P-Loss observes drastic changes due to intervening in similar relationships. Hence, we argue that P-Loss can intervene in similar relationships sparsely. In other words, it eliminates confounder $S$ without affecting other confounders. As a result, the model trained with P-Loss can be roughly formulated as a disentangled factorization.

\textit {Disentangled claim in Equation (\ref{eq:disentangled-XY}).} This subsection designs a new metric, \ie, mean Correction Rate (mC@K), to support our disentangled claims. mC@K calculates the balanced fraction of rates that the false relationships are corrected, where K shares the same meaning as in mainstream metrics (\eg, R@K and mR@K). In keeping with the claim to be demonstrated, Fig.~\ref{dis_longtail} shows mC@K on the tail categories, which allows us to evaluate the performance of the AL-Adjustment in alleviating the long-tailed distribution problem. These results clearly show that AL-Adjustment can adjust a considerable number of false predictions of tail categories to correct ones, and, hence, the long-tailed distribution confounder can be removed by our proposed adjustment procedure. Taking together the disentangled claim in Equation (\ref{eq:disentangled-XYL}), we can naturally come to the disentangled claim in Equation (\ref{eq:disentangled-XY}).

\section{Conclusion}
In this paper, we have proposed a novel causal modeling framework, TsCM, for unbiased scene graph generation, which decouples the causal intervention into two stages to eliminate semantic confusion bias and long-tailed distribution bias, where the former bias is rarely explored in the existing debiasing methods. In stage 1, we analyzed that the SCM modeled for SGG is always causal-insufficient and the sparsity of relationship categories. On this basis, a causal representation learning method is proposed to achieve sparse interventions on semantic confusion bias in the case of insufficient causality. As a result, this stage also provides a disentangled factorization. Benefiting from this factorization, stage 2 then proposes causal calibration learning to intervene sparsely and independently in the long-tailed distribution bias to achieve unbiased predictions. Experiments were conducted on the popular SGG backbones and dataset, and our method achieved state-of-the-art debiasing performance. Furthermore, our method achieved a better tradeoff between recall rate and mean recall rate thanks to the sparse causal interventions.

Although our method can remove multiple biases in the SGG task, it is still challenging to overcome the unobservable biases. In the future, we will focus on exploring the unobservable biases and develop the automatic debiasing causal framework to pursue unbiased SGG predictions.

\ifCLASSOPTIONcaptionsoff
  \newpage
\fi

\bibliographystyle{IEEEtran}
\bibliography{biblio}

\end{document}